\documentclass{article}

\usepackage{amsmath,amsfonts,bm}

\def\eqref#1{equation~\ref{#1}}

\def\1{\bm{1}}

\DeclareMathAlphabet{\mathsfit}{\encodingdefault}{\sfdefault}{m}{sl}
\SetMathAlphabet{\mathsfit}{bold}{\encodingdefault}{\sfdefault}{bx}{n}

\DeclareMathOperator*{\argmax}{arg\,max}
\DeclareMathOperator*{\argmin}{arg\,min}

\usepackage{microtype}
\usepackage{graphicx}
\usepackage{subcaption}
\usepackage{booktabs} %

\usepackage[hyphens]{url}
\usepackage{hyperref}
\hypersetup{breaklinks=true}

\usepackage{float}
\usepackage{algorithm}
\usepackage{algorithmic}

\usepackage[accepted]{icml2026}

\usepackage[most]{tcolorbox}

\usepackage{amssymb}
\usepackage{mathtools}
\usepackage{amsthm}

\usepackage{amssymb}%
\usepackage{pifont}%

\usepackage{booktabs}
\usepackage{multirow}
\usepackage[inline]{enumitem}
\usepackage{graphicx}
\usepackage{booktabs}
\usepackage{caption}
\usepackage{amsmath}
\usepackage{tabularx}   %
\usepackage{array}      %
\usepackage{makecell}   %

\usepackage{xparse}

\makeatletter
\newcommand{\appendixtableofcontents}{%
  \section*{Contents of Appendix}%
  \setcounter{tocdepth}{2}%
  \@starttoc{atoc}%
}

\newcommand{\apx@writetocline}[4]{%
  \begingroup
    \let\label\@gobble
    \let\index\@gobble
    \let\glossary\@gobble
    \ifdefined\hyper@anchor
      \protected@write\tf@atoc{}{%
        \string\contentsline{#1}{#2}{#3}{#4}%
      }%
    \else
      \protected@write\tf@atoc{}{%
        \string\contentsline{#1}{#2}{#3}%
      }%
    \fi
  \endgroup
}

\hypersetup{colorlinks=true, linkcolor=blue, citecolor=blue, urlcolor=blue}

\newcommand{\activeuf}{\textsc{ActiveUltraFeedback}}

\usepackage[capitalize,noabbrev]{cleveref}

\theoremstyle{plain}

\theoremstyle{definition}

\theoremstyle{remark}

\usepackage[textsize=tiny]{todonotes}

\icmltitlerunning{ActiveUltraFeedback: Efficient Preference Data Generation using Active Learning}

\begin{document}

\twocolumn[
  \icmltitle{\texorpdfstring{ActiveUltraFeedback: \\ Efficient Preference Data Generation using Active Learning}{ActiveUltraFeedback: Efficient Preference Data Generation using Active Learning}}

  \icmlsetsymbol{equal}{*}

    \begin{icmlauthorlist}
    \icmlauthor{Davit Melikidze}{eth,equal}
    \icmlauthor{Marian Schneider}{eth,equal}
    \icmlauthor{Jessica Lam}{ini,equal}
    \icmlauthor{Martin Wertich}{eth,equal}{\\[0.15em]}
    \icmlauthor{Ido Hakimi}{eth,ethai}
    \icmlauthor{Barna Pásztor}{eth,ethai}
    \icmlauthor{Andreas Krause}{eth,ethai}
    \end{icmlauthorlist}
    
    \icmlaffiliation{eth}{ETH Zurich}
    \icmlaffiliation{ini}{Institute of Neuroinformatics, University of Zurich and ETH Zurich}
    \icmlaffiliation{ethai}{ETH AI Center}
    
    \icmlcorrespondingauthor{Davit Melikidze}{dmelikidze@ethz.ch}
    \icmlcorrespondingauthor{Marian Schneider}{smarian@ethz.ch}
    \icmlcorrespondingauthor{Jessica Lam}{jehong@ethz.ch}
    \icmlcorrespondingauthor{Martin Wertich}{mwertich@ethz.ch}

  \icmlkeywords{Machine Learning, LLMs, Active Learning, Preference Optimization}

  \vskip 0.15in
]

\printAffiliationsAndNotice{\icmlEqualContribution}

\begin{abstract}
Reinforcement Learning from Human Feedback (RLHF) has become the standard for aligning Large Language Models (LLMs), yet its efficacy is bottlenecked by the high cost of acquiring preference data, especially in low-resource and expert domains.
To address this, we introduce \activeuf, a modular active learning pipeline that leverages uncertainty estimates to dynamically identify the most informative responses for annotation.
Our pipeline facilitates the systematic evaluation of standard response selection methods alongside \textsc{Double Reverse Thompson Sampling (DRTS)} and \textsc{DeltaUCB}, two novel methods prioritizing response pairs with large predicted quality gaps, leveraging recent results showing that such pairs provide good signals for fine-tuning.
Our experiments demonstrate that \activeuf{} yields high-quality datasets that lead to significant improvements in downstream performance, notably achieving comparable or superior results with as little as one-sixth of the annotated data relative to static baselines.
Our pipeline is available at \url{https://github.com/lasgroup/ActiveUltraFeedback} and our preference datasets at \url{https://huggingface.co/ActiveUltraFeedback}.
\end{abstract}

\section{Introduction}

\begin{figure}[t]
    \centering
    \includegraphics{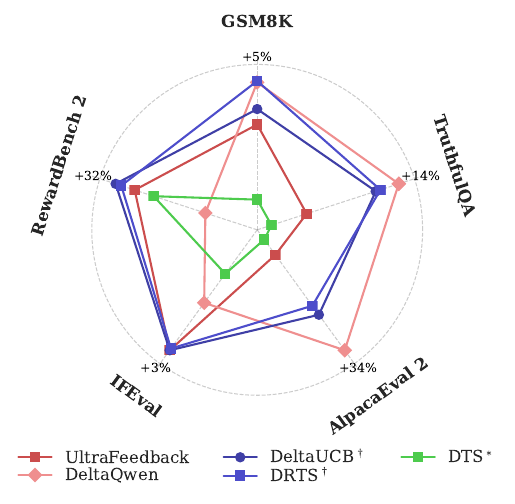}
    \caption{Comparison of response pair selection methods on downstream and reward model benchmarks deployed in \activeuf{}. The scores have been averaged over four datasets (see \cref{sec:input_prompt_dataset_ablation}) of different scales, and indicate improvement over the base model. * denotes an existing dueling bandit method and \textdagger{} indicates our novel active delta learning methods.
    }
    \label{fig:teaser}
\end{figure}

\looseness=-1
Reinforcement Learning from Human Feedback (RLHF) has established itself as a critical methodology to align Large Language Models (LLMs) with human preferences~\citep{ziegler2019fine, ouyang2022training}.
RLHF guides the model using human feedback articulated as pairwise preferences over potential outputs, resulting in more naturalistic and human-like behaviour~\citep{christiano2017deep}. The standard implementation involves training a reward model, followed by model optimization with Proximal Policy Optimization (PPO)~\citep{schulman2017proximal} to maximize expected rewards~\citep{ouyang2022training}. Alternatively, Direct Preference Optimization (DPO)~\citep{rafailov2023direct} circumvents the need for a separate reward model by optimizing the model directly on the dataset of pairwise preferences.
The potential efficacy of these methods increases with the quality of the preference data, but human annotation is expensive to obtain, especially in low-resource or expert domains. Consequently, a promising direction for low-cost and scalable preference dataset creation is to reduce annotation requirements by identifying and labelling only the most informative response pairs.

\looseness=-1
Existing works such as UltraFeedback~\citep{cui2024ultrafeedbackboostinglanguagemodels}, Magpie~\citep{xu2024magpiealignmentdatasynthesis}, and Nectar~\citep{starling2023} generate response pairs through static, passive heuristics.
Common choices are random or best-of-$N$ sampling~\citep{cui2024ultrafeedbackboostinglanguagemodels, starling2023}, which are either inefficient or require multiple annotations per prompt. Our experiments show that neither results in high-quality datasets.
More recently, the Delta Learning Hypothesis (DLH)~\citep{geng2025deltalearninghypothesispreference} proposed a novel approach by pairing models of different sizes within a single family (e.g., small vs. large) to form contrastive pairs without annotation. While effective for common applications, this rigidity limits DLH to domains within the chosen model family's training data, and as our experiments show, its performance is limited to DPO fine-tuning.
Therefore, the question of collecting high-quality preference datasets not tied to specific algorithms while keeping the need for costly annotation low remains open.

\looseness=-1
In this work, we propose \activeuf, a modular preference data collection pipeline.
Our framework is modeled after the contextual dueling bandit problem ~\citep{dudik2015contextualduelingbandits}. In this setup, the prompt serves as the context, and the objective is to select two candidate responses (the arms) from a diverse pool for annotation.
We maintain a probabilistic estimate of response quality, updated sequentially as data is collected, to guide the selection of subsequent pairs. Within this framework, we conduct a systematic evaluation of response pair selection methods, comparing standard dueling bandit approaches
against established heuristics.
Furthermore, we introduce \textsc{Double Reverse Thompson Sampling (DRTS)} and \textsc{DeltaUCB}, two novel methods integrating the insights of the Delta Learning Hypothesis~\citep{geng2025deltalearninghypothesispreference}  by prioritizing pairs with high predicted quality gaps rather than simply minimizing regret.
As previewed in \cref{fig:teaser}, \activeuf{} with \textsc{DRTS} and \textsc{DeltaUCB} consistently outperforms prior heuristics and standard dueling-bandit baselines across both fine-tuned and reward-model benchmarks.
Notably, \activeuf{} demonstrates strong sample-efficiency, matching or outperforming previous methods using only one-sixth of the data, requiring only a single pairwise comparison per prompt for annotation, and not being confined to a single model family. This efficiency enables its application to domains not supported by previous methods. Our detailed ablations demonstrate that these results hold across various datasets and fine-tuning algorithms.

In summary, our contributions are as follows:
\begin{itemize}
    \looseness=-1
    \item We introduce \activeuf, a modular preference data generation pipeline, that can be deployed with any response selection and uncertainty quantification methods to guide data collection.
    \item We are the first to perform a systematic comparison of dueling bandit acquisition functions and common data collection heuristics across a comprehensive evaluation suite covering both reward modeling and diverse downstream benchmarks.
    \item We introduce two novel response pair selection approaches, \textsc{DRTS} and \textsc{DeltaUCB}, that generate datasets yielding strong performance across prompt sources, tasks, and fine-tuning algorithms, while relying on fewer annotations.
    \item We open-source \activeuf{} to allow for easy adoption in existing data pipelines and release artifacts, such as datasets and models.
\end{itemize}

\section{Related Work}\label{sec:related_work}

\begin{figure*}[htb]
    \centering
    \includegraphics[width=0.9\linewidth]{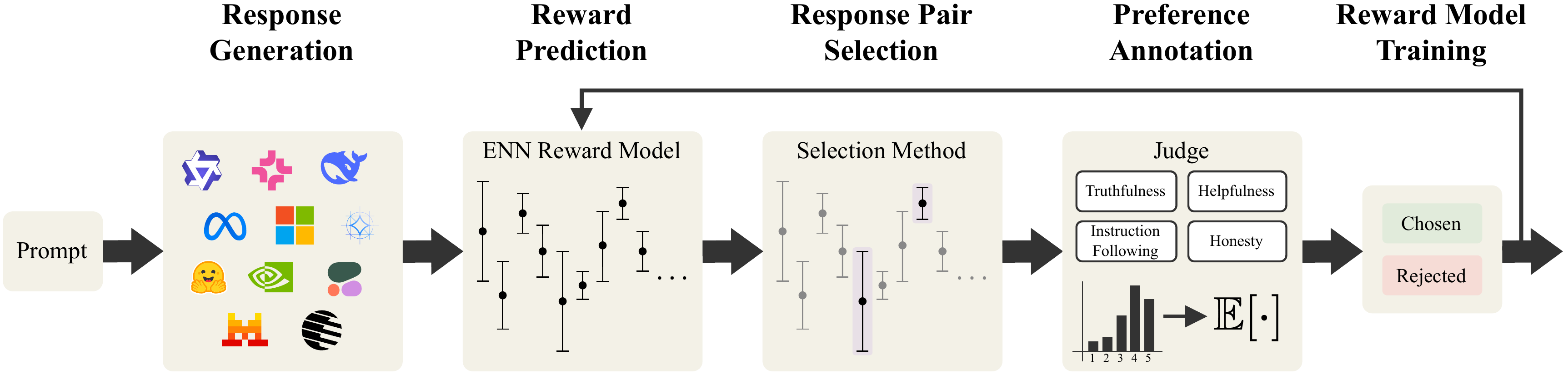}
    \caption{The \activeuf{} pipeline. For each prompt, responses are generated from a large pool of LLMs, the rewards for the responses are predicted with corresponding uncertainties, and a pair of responses is selected for preference annotation. Each new batch of preference data is used to train the reward model, improving the accuracy of reward and uncertainty estimates for subsequent iterations. The displayed procedure is performed in a looping manner until all prompts have been processed.}
    \label{fig:pipeline_overview}
\end{figure*}

\looseness=-1
Reinforcement Learning from Human Feedback (RLHF) is a common method for training models on qualitative objectives concerning human preferences~\citep{christiano2017deep, ziegler2019fine, ouyang2022training}.
A standard pipeline involves training a reward model on pairwise comparison data, then reinforcement learning algorithms like PPO~\citep{schulman2017proximal} optimize the model. Alternatively, Direct Preference Optimization (DPO)~\citep{rafailov2023direct} offers a solution that combines the two steps.
However, the efficacy of these methods is bottlenecked by data provenance.
Traditional pipelines rely on manual annotation~\citep{ziegler2019fine, stiennon2022learningsummarizehumanfeedback, bai2022traininghelpfulharmlessassistant} or noisy indirect signals~\citep{pmlr-v162-ethayarajh22a}. The former is prohibitively expensive to scale, while the latter lacks control over domain coverage and data quality.

\looseness=-1
To scale up supervision and leverage the performance of frontier models, recent efforts, such as UltraFeedback~\citep{cui2024ultrafeedbackboostinglanguagemodels}, Magpie~\citep{xu2024magpiealignmentdatasynthesis}, and Nectar~\citep{starling2023} have shifted towards generating synthetic data. They follow a common paradigm: a pool of instruction-tuned LLMs generates multiple candidate responses per prompt, then the candidates are scored or ranked~\citep{starling2023} by a judge, and a chosen-rejected pair is selected~\citep{cui2024ultrafeedbackboostinglanguagemodels, wang2024interpretable}.
While these methods have successfully trained open-source models like Zephyr~\citep{tunstall2024zephyr}, Tulu 3~\citep{lambert2025tulu3pushingfrontiers}, and Olmo 2~\citep{olmo20252olmo2furious}, they apply the same selection strategy to every prompt regardless of response quality uncertainty. This lack of adaptivity often results in sample inefficiency and low-quality datasets, as the system consumes budget on trivial comparisons while missing high-information pairs.
Alternatively, the Delta Learning Hypothesis (DLH)~\citep{geng2025deltalearninghypothesispreference} employs a structural heuristic, pairing models of different sizes (e.g., 0.6B vs. 32B) within a single family to guarantee a quality gap without requiring a judge. Despite its success in training Olmo 3~\citep{olmo2025olmo} and SmolLM3~\citep{bakouch2025smollm3}, DLH is rigidly confined to intra-family comparisons, limiting its applicability to their often unknown training domains.

Recent works address sample inefficiency in RLHF by formulating it as a contextual duelling bandit problem~\citep{dudik2011efficientoptimallearningcontextual}. 
For reward model training, prior work adapts Double Thompson Sampling (DTS)~\citep{dwaracherla2024efficientexplorationllms}, applies information-theoretic selection~\citep{shen2025reviving}, and uses uncertainty to estimate preference quality and adaptively weight samples~\citep{zhang-etal-2025-dorm}. 
For model fine-tuning, uncertainty estimates over predicted rewards improve sample efficiency through uncertainty-based data selection~\citep{liu2024sampleefficientalignmentllms,muldrew2024activepreferencelearninglarge,mehta2025sampleefficientpreferencealignment,cercola2025efficientreinforcementlearninghuman}, exploration bonuses~\citep{liang2022rewarduncertaintyexplorationpreferencebased}, or uncertainty-regularized objectives that penalize high-uncertainty rewards during RL optimization~\citep{zhai2023uncertaintypenalizedreinforcementlearninghuman}. 
However, the literature remains fragmented: studies typically focus narrowly on either reward model training~\citep{dwaracherla2024efficientexplorationllms,shen2025reviving,zhang-etal-2025-dorm} or policy optimization~\citep{muldrew2024activepreferencelearninglarge,liu2024sampleefficientalignmentllms,kveton2025activelearningdirectpreference,mehta2025sampleefficientpreferencealignment}, often within a single model family.
In contrast, we do not restrict our scope to a single selection method, application, or optimization algorithm.

We bridge this gap by proposing a unified, modular pipeline that enables evaluating response pair selection strategies across both downstream fine-tuning and reward modeling. Within this framework, we benchmark active learning strategies directly against static heuristics and introduce novel methods that operationalize insights from the Delta Learning Hypothesis. Our pipeline generates high-quality datasets for both reward modeling and model fine-tuning, and performs well with multiple preference optimization algorithms.

\section{Background} \label{sec:background}

Reinforcement Learning from Human Feedback (RLHF) aligns models with human intent by learning from a dataset of pairwise comparisons $\mathcal{D} = \{(x_i, y_i^+, y_i^-)\}_{i=1}^N$, where $x_i$ denotes a prompt and $(y_i^+, y_i^-)$ denotes candidate responses with $y_i^+$ preferred to $y_i^-$. For brevity, we drop the indexing by $i$ for this section.
The standard approach~\citep{christiano2017deep} proceeds in two stages. First, a reward model $r_\phi(x, y)$ is trained to approximate the latent human preference distribution. This typically relies on the Bradley-Terry model~\citep{bradley1952rank}, which assumes that the comparison feedback is drawn from a Bernoulli distribution and the probability of $y^+$ being preferred to $y^-$ is given by the sigmoid of their reward difference, i.e.,
\begin{equation}
    \label{eq:bradley_terry}
    p(y^+ \succ y^- \mid x) = \operatorname{s}(r(x, y^+) - r(x, y^-)),
\end{equation}
where $\operatorname{s}(x) = (1+e^{-x})^{-1}$ is the sigmoid function and $r$ is an unknown latent scalar function.
The parametrized reward model $r_\phi$ is then optimized to estimate the unknown reward function $r$ by minimizing the negative log-likelihood of the dataset in $\mathcal{D}$.
Second, the model, $\pi_\theta$, is optimized to maximize the regularized objective
\begin{equation}
    \label{eq:preference_optimization_objective}
    \mathcal{J}(\theta) = \mathbb{E}_{x \sim \mathcal{D}, y \sim \pi_\theta(\cdot|x)} \left[ r_\phi(x, y) - \lambda \operatorname{KL}(\pi_\theta \| \pi_{\text{ref}}) \right],
\end{equation}
where $\operatorname{KL}$ denotes the Kullback-Leibler divergence from a reference model $\pi_{\text{ref}}$ and $\lambda$ controls the strength of the regularization.
Direct Preference Optimization (DPO)~\citep{rafailov2023direct} is a widely used alternative that improves computational efficiency by combining the reward modeling and policy fine-tuning steps, turning RLHF into a supervised learning task.
Regardless of the optimisation approach, standard RLHF methods consider $\mathcal{D}$ as a fixed, static artifact.

While the standard RLHF approaches only use a pointwise estimate for the reward function $r_\phi$, we leverage uncertainty estimates to guide data collection.
Let $\underline{r}_\phi(x, y)$ and $\overline{r}_\phi(x, y)$ denote the lower and upper confidence bounds of the reward estimate. Under the Bradley-Terry assumption, the upper confidence bound (UCB) probability $\overline{p}$ that a response $y_{j}$ is preferred over another response $y_{j'}$ is defined as
\begin{equation}
    \label{eq:ucb}
    \overline{p}_\phi(y_{j} \succ y_{j'}) = \operatorname{s}(\overline{r}_\phi(x, y_{j}) - \underline{r}_\phi(x, y_{j'})).
\end{equation}
Conversely, the lower confidence bound (LCB) probability $\underline{p}$ is defined by the worst-case reward difference
\begin{equation}
    \label{eq:lcb}
    \underline{p}_\phi(y_{j} \succ y_{j'}) = \operatorname{s}(\underline{r}_\phi(x, y_{j}) - \overline{r}_\phi(x, y_{j'})).
\end{equation}
These probabilistic bounds serve as the foundation for response selection methods described in \cref{sec:response_pair_acquisition}.

\section{The \activeuf{} Pipeline} \label{sec:pipeline}

In this section, we introduce \activeuf, our scalable and modular pipeline for creating high-quality preference datasets without extensive annotation requirements. Given a set of $N$ prompts, $\mathcal{P}=\{x_i\}_{i=1}^N$, \activeuf{} starts with an empty dataset $\mathcal{D} = \emptyset$, processes the prompts in $\mathcal{P}$ iteratively in batches, and appends the new data points to $\mathcal{D}$. Unlike prior work, we present a unified active learning pipeline for preference data generation. \textsc{ActiveUltraFeedback} supports plug-and-play uncertainty-aware acquisition functions, evaluates them across reward modeling and downstream preference tuning, and introduces two new delta-oriented methods, \textsc{DRTS} and \textsc{DeltaUCB}. The five key steps for each batch, illustrated in \cref{fig:pipeline_overview}, are as follows:

\begin{enumerate}
    \item \textbf{Response Generation}: For each prompt $x_i$ in the batch, generate a diverse set of candidate responses $\{y_{i,j}\}_{j=1}^m$ from a pool of $m$ LLMs (\cref{sec:response_generation}).
    \item \textbf{Reward Prediction}: For each prompt--response pair $(x_i, y_{i,j})$, estimate $\underline{r}_\phi(x_i, y_{i,j})$ and $\overline{r}_\phi(x_i, y_{i,j})$ (\cref{sec:reward_prediction}).
    \item \textbf{Response Pair Selection}: Select two responses $(y_{i,j}, y_{i,j'})$ for each prompt in the batch for pairwise comparison (\cref{sec:response_pair_acquisition}).
    \item \textbf{Preference Annotation}: Collect preference annotations and append the resulting triplets, $(x_i, y_i^+, y_i^-)$, to $\mathcal{D}$ (\cref{sec:oracle_preference_annotation}).
    \item \textbf{Reward Model Training}: Update the reward model's parameters, $\phi$, with the dataset $\mathcal{D}$ collected thus far (\cref{sec:reward_model_training}).
\end{enumerate}

\subsection{Response Generation} \label{sec:response_generation}

Given an input prompt $x_i$, we employ a model pool of $m$ LLMs to generate candidate responses $\{y_{i,j}\}_{j=1}^m$. Our model pool comprises $m = 30$ open-weight LLMs from 12 families, including Qwen 2.5~\citep{qwen2025qwen25technicalreport}, Qwen 3~\citep{yang2025qwen3technicalreport}, Llama 3~\citep{grattafiori2024llama3herdmodels}, Gemma 3~\citep{gemmateam2024gemmaopenmodelsbased}, and SmolLM 2~\citep{allal2025smollm2smolgoesbig}. Following the UltraFeedback pipeline's approach~\citep{cui2024ultrafeedbackboostinglanguagemodels, lambert2025tulu3pushingfrontiers, olmo20252olmo2furious}, for each prompt--LLM pair, we select a guiding principle (from ``helpfulness'', ``truthfulness'', and ``honesty'') at random to create more diverse responses.

The combination of aspects and the diverse model pool ensures that the candidate responses provide a broad content and quality diversity for the response pair selection methods. We defer further details on the model pool (\cref{tab:response_model_pool}), principles (\cref{app:response_principles}), and the used prompt templates (\cref{app:response_generation_prompt_templates}) to the Appendix.

\subsection{Reward Prediction} \label{sec:reward_prediction}

To operationalize the uncertainty estimates defined in \cref{sec:background}, we employ the Epistemic Neural Network (ENN) framework~\citep{osband2023epistemicneuralnetworks}.
Following prior works for active learning in RLHF~\citep{dwaracherla2024efficientexplorationllms, melo2024deep, liu2024sampleefficientalignmentllms}, we implement the ENN as an ensemble of shallow Multi-Layer Perceptrons with a shared, frozen backbone, deriving the final reward $r_\phi(x_i, y_j)$ as the ensemble mean and uncertainty $\sigma_\phi(x_i, y_j)$ as the standard deviation. These quantities define the upper and lower confidence bounds for the reward estimate 
\begin{align*}
    \overline{r}_\phi(x_i, y_j) &= r_\phi(x_i, y_j) + \beta \sigma_\phi(x_i, y_j), \\
    \underline{r}_\phi(x_i, y_j) &= r_\phi(x_i, y_j) - \beta \sigma_\phi(x_i, y_j)
\end{align*}
respectively, where $\beta > 0$ is a scaling parameter, as well as the UCB $\overline{p}_\phi$ (\cref{eq:ucb}) and LCB $\underline{p}_\phi$ (\cref{eq:lcb}) for comparisons between response pairs.
Additional details on the network architecture are provided in \cref{app:reward_model_architecture}.

\subsection{Response Pair Selection}\label{sec:response_pair_acquisition}

For each prompt $x_i$, we select a response pair $(y_{i,j}, y_{{i,j'}})$ for preference annotation using a response pair selection method. We explore four baseline heuristics that do not make use of the reward estimates and three methods developed for the Dueling Bandit problem~\citep{bengs2021preference}.
Additionally, we propose two novel methods, \textsc{DRTS} and \textsc{DeltaUCB}, based on the Delta Learning Hypothesis (DLH)~\citep{geng2025deltalearninghypothesispreference}.
We provide an overview of the algorithms here and defer further details to \cref{app:details_on_acquisition_functions}.

\begin{table}[t]
    \centering
    \caption{Overview of response pair selection methods and the number of responses that need to be annotated per prompt. \textdagger{} indicates the methods that we propose.}
    \begin{tabular}{lc}
        \toprule
        \multirow{2}{*}{\textbf{Methods}} 
            & \multirow{2}{*}{\shortstack[c]{\textbf{\# Responses}\\\textbf{to Annotate}}} \\ \\
        \midrule
        \textit{Baseline Heuristics} \\
            \, \textsc{Random}        & 2   \\
            \, \textsc{MaxMin}        & $m$ \\
            \, \textsc{UltraFeedback}~\citep{cui2024ultrafeedbackboostinglanguagemodels} & 4   \\
            \, \textsc{DeltaQwen}~\citep{geng2025deltalearninghypothesispreference} & 0   \\
        \midrule
        \textit{Dueling Bandit Methods} \\
            \, \textsc{InfoMax}~\citep{infobasedactiveselection} & 2 \\
            \, \textsc{DTS}~\citep{wu2016doublethompsonsamplingdueling}                   & 2 \\
            \, \textsc{MaxMinLCB}~\citep{pasztor2024bandits} & 2 \\
        \midrule
        \textit{Active Delta Learning Methods} \\
            \, \textsc{DRTS}\textsuperscript{\textdagger}     & 2 \\
            \, \textsc{DeltaUCB}\textsuperscript{\textdagger} & 2 \\
        \bottomrule
    \end{tabular}
    \label{tab:acquisition_functions}
\end{table}

\begin{table*}[ht]
    \centering
    \caption{Comparison between all response pair selection methods, based on the reward model and fine-tuned model (DPO) performance after training the same base model on each generated dataset. The base model score is given for reference, and all scores are reported as relative deltas to it, with higher values indicating better performance. We also provide the deltas achieved with the original response pairs in \textbf{UltraFeedback}. $\dagger$ denotes our proposed methods. Best scores are marked in bold.}
    \begin{tabular}{l|ccccc|c}
        \toprule
        \textbf{Method} & \textbf{GSM8K} & \textbf{IFEval} & \textbf{TruthfulQA} & \textbf{AlpacaEval 2} &\textbf{Mean} & \textbf{RewardBench 2} \\
        \midrule
        Base Model     & \phantom{+}0.758 & \phantom{+}0.713 & \phantom{+}0.468 & \phantom{+}0.083 & \phantom{+}0.506 & \phantom{+}0.290 \\
        Original       &           +0.039 &           +0.025 &           +0.055 &           +0.030 &           +0.037 &           +0.295 \\
        \midrule
        \textsc{Random}         & +0.024 & +0.028 & +0.056 & +0.077 &         +0.046  &         +0.278  \\
        \textsc{UltraFeedback}  & +0.037 & -0.001 & +0.039 & +0.072 &         +0.036  &         +0.287  \\
        \textsc{MaxMin}         & +0.022 & -0.016 & \textbf{+0.150} & +0.289 &         +0.111  & +0.318 \\
        \textsc{DeltaQwen}      & \textbf{+0.055} & +0.047 & +0.130 & \textbf{+0.316} & \textbf{+0.137} &         +0.100  \\
        \midrule
        \textsc{InfoMax}        & +0.011 & +0.019 & +0.018 & +0.020 &         +0.016  & +0.297 \\ %
        \textsc{DTS}            & +0.011 & +0.034 & +0.013 & +0.037 & +0.023 &         +0.224  \\ %
        \textsc{MaxMinLCB}      & +0.015 & +0.017 & +0.006 & +0.027 &         +0.016  &         +0.230  \\ %
        \midrule 
        \textsc{DRTS}$^\dagger$           & \textbf{+0.055} & \textbf{+0.050} & +0.143 & +0.259 & +0.127 &         +0.312  \\ %
        \textsc{DeltaUCB}$^\dagger$       & +0.040 & +0.025 & +0.137 & +0.281 &         +0.120  & \textbf{+0.339} \\ %
        \bottomrule
    \end{tabular}
    \label{tab:acquisition_function_comparison}
\end{table*}

\paragraph{Baseline Heuristics} We evaluate four passive baseline heuristics that operate independently of reward estimates.
\begin{enumerate*}[label=(\roman*)]
    \item \textsc{Random} samples a pair uniformly at random from the candidate set; 
    \item \textsc{MaxMin} queries a judge for the entire candidate set to identify the responses with the highest and lowest quality;
    \item\textsc{UltraFeedback}~\citep{cui2024ultrafeedbackboostinglanguagemodels} samples four responses uniformly at random, queries a judge on their quality, and returns the highest-scoring one as the preferred response paired with a randomly selected one from the remaining three; 
    \item \textsc{DeltaQwen}~\citep{geng2025deltalearninghypothesispreference} selects the responses generated by the Qwen 3 0.6B and 32B models, with the latter considered as the preferred response. 
\end{enumerate*}

\begin{figure*}[t]
    \centering
    \includegraphics{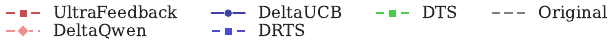}

    \begin{subfigure}[b]{0.5\textwidth}
        \centering
        \includegraphics[trim={0 0 0 0}, clip]{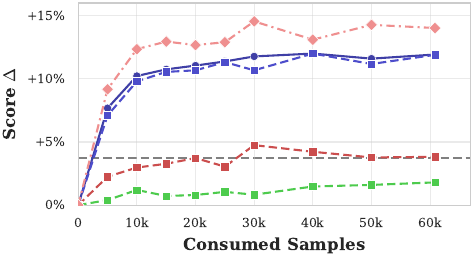}
        \caption{Fine-tuned Models}
        \label{fig:sample_efficiency_dpo}
    \end{subfigure}%
    \hfill 
    \begin{subfigure}[b]{0.5\textwidth}
        \centering
        \includegraphics[trim={0 0 0 0}, clip]{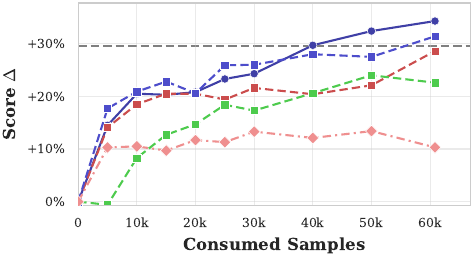}
        \caption{Reward Models}
        \label{fig:sample_efficiency_rm}
    \end{subfigure}
    
    \caption{Mean performance trajectories for fine-tuned and reward models as a function of consumed samples on the \activeuf{} prompt pool using DPO. All curves share the same prompts and differ only in the response pair selection strategy. For readability, we subset the selection strategies; for a full comparison, see~\cref{fig:full_sample_efficiency} in the Appendix. We compare datasets generated via \textsc{ActiveUltraFeedback} using various selection methods, and also report the score achieved by the original \textbf{UltraFeedback} dataset \citep{cui2024ultrafeedbackboostinglanguagemodels} with its original response pairs.}
    \label{fig:sample_efficiency} 
\end{figure*}

\paragraph{Dueling Bandit Methods} We adopt three acquisition functions from prior literature on dueling bandits:
\begin{enumerate*}[label=(\roman*)]
    \item \textsc{InfoMax}~\citep{infobasedactiveselection} prioritizes pure exploration by selecting the response pair with the highest joint uncertainty, regardless of the predicted reward quality: $\operatorname*{arg\,max}_{j \neq j'} \overline{p}_\phi(y_{i,j} \succ y_{i,j'}) - \underline{p}_\phi(y_{i,j} \succ y_{i,j'})$;
    \item \textsc{Double Thompson Sampling (DTS)}~\citep{wu2016doublethompsonsamplingdueling} addresses the exploration-exploitation trade-off by drawing two independent samples from the reward posterior and selecting the responses that maximize them;
    \item \textsc{MaxMinLCB}~\citep{pasztor2024bandits} considers the pairwise LCB (\cref{eq:lcb}) and selects the pair $(j_1, j_2)$ where $j_1 = \argmax_j \min_{j' \neq j} \underline{p}_\phi(y_j \succ y_{j'})$ maximizes the minimum LCB against any other response, and $j_2 = \argmin_{j \neq j_1} \underline{p}_\phi(y_{j_1} \succ y_{j})$ minimizes the LCB against $j_1$.
\end{enumerate*}
These algorithms offer no-regret guarantees (\textsc{DTS}, \textsc{MaxMinLCB}) or sample complexity bounds for identifying the optimal response (\textsc{InfoMax}).

\paragraph{Active Delta Learning Methods} We introduce two novel methods based on the Delta Learning Hypothesis~\citep{geng2025deltalearninghypothesispreference}, which states that the absolute quality of the responses is less important than the relative difference, and proposed the \textsc{DeltaQwen} method introduced above.

\textsc{Double Reversed Thompson Sampling (\textsc{DRTS})} selects one response that maximizes and another that minimizes their respective samples from the reward posterior. This strategy explicitly targets pairs with a significant delta in quality, while the underlying stochastic sampling preserves exploration and diversity.

\textsc{DeltaUCB} identifies pairs with the largest optimistic quality difference by selecting the pair $(y_{i,j}, y_{i,j'})$ that maximizes the probability that $j$ is preferred over $j'$ in the best-case scenario: $\operatorname*{arg\,max}_{j \neq j'} \overline{p}_\phi(y_{i,j} \succ y_{i,j'})$. By relying on these optimistic bounds, \textsc{DeltaUCB} guides exploration toward pairs that plausibly exhibit significant quality differences, without requiring stochastic sampling.

\subsection{Preference Annotation}\label{sec:oracle_preference_annotation}
After the response pairs $(y_{i,j}, y_{i,j'})$ for each prompt $x_i$ are selected, we query a judge for the pairwise comparison feedback and, following the annotation, append $(x_i, y_i^+, y_i^-)$ to the dataset $\mathcal{D}$. To facilitate scalable and reproducible experiments, we employ a large LLM instead of human annotators. Our goal is not to fully replace human preference judgments, but to enable controlled large-scale comparisons of response pair selection methods while reducing annotation costs. Specifically, a judge LLM independently scores each response on a 1--5 Likert scale across four quality aspects: truthfulness, instruction following, honesty, and helpfulness. The response with the highest average score is then selected as preferred.
To ensure high-quality labels, we validated our annotation setup through extensive experiments comparing different judges, prompting strategies, and scoring mechanisms. Further details are provided in \cref{app:annotation}.

\subsection{Reward Model Training}\label{sec:reward_model_training}
Finally, we update the ENN model to improve its reward estimates using the latest batch of preference data combined with previously collected samples. For details on hyperparameters and the training procedure, see \cref{app:reward_model_training}.

\section{Evaluation}

In this section, we evaluate the response pair selection methods  (\cref{sec:response_pair_acquisition}) deployed in \activeuf{} by investigating the following research questions:

\begin{enumerate}
    \item \textbf{Performance}: Can \activeuf{} generate high-quality datasets (\cref{sec:acquisition_evaluation}), and which response pair selection method achieves the best performance?
    \item \textbf{Efficiency}: Does active response pair selection provide sample efficiency improvements (\cref{sec:sample_efficiency_evaluation}), yielding equal or higher scores using fewer annotated samples?
    \item \textbf{Generalization}: Do results generalize across prompt datasets (\cref{sec:input_prompt_dataset_ablation}) and preference optimization algorithms (\cref{sec:preference_optimization_algorithm_ablation})?
\end{enumerate}

\subsection{Implementation Details} \label{sec:implementation_details}

\paragraph{Datasets} We choose the \textbf{UltraFeedback} dataset\footnote{\href{https://huggingface.co/datasets/allenai/ultrafeedback_binarized_cleaned}{allenai/ultrafeedback\_binarized\_cleaned}}~\citep{cui2024ultrafeedbackboostinglanguagemodels} as our primary set of prompts $\mathcal{P}$ and consider further prompt collections in \cref{sec:input_prompt_dataset_ablation}.

\looseness=-1
\paragraph{Evaluation} To evaluate the datasets collected by \activeuf, we consider the two steps of RLHF described in \cref{sec:background}, reward model training and model fine-tuning, separately.
First, we train a standard reward model using the negative log likelihood minimization of the Bradley-Terry model defined in \cref{eq:bradley_terry} and evaluate it on the RewardBench 2 benchmark~\citep{malik2025rewardbench2advancingreward}.
To keep our evaluation protocol standardized, we train the reward model independently of the ENN described in \cref{sec:reward_prediction}.
To isolate reward modeling and preference fine-tuning, we use DPO~\citep{rafailov2023direct}, which combines the two steps of RLHF. We evaluate other direct optimization algorithms in \cref{sec:preference_optimization_algorithm_ablation}.
The fine-tuned models are then evaluated on the 
GSM8K~\citep{cobbe2021training}, 
IFEval~\citep{zhou2023instructionfollowingevaluationlargelanguage}, TruthfulQA~\citep{lin2022truthfulqa}, and AlpacaEval 2~\citep{dubois2024length} benchmarks covering the crucial capabilities of mathematical reasoning, instruction-following, knowledge recall, and human preference.
Both trainings for evaluation are initialized from the Tulu 3 8B SFT model\footnote{\href{https://huggingface.co/allenai/Llama-3.1-Tulu-3-8B-SFT}{allenai/Llama-3.1-Tulu-3-8B-SFT}}~\citep{lambert2025tulu3pushingfrontiers} and all scores are reported as deltas relative to the base model.
We measured our results' sensitivity to the inherent stochastic nature of our pipeline and consider a difference of at least $0.008$ for the downstream benchmarks and $0.02$ for RewardBench 2 to be significant.
Detailed analysis is provided in \cref{app:training_stability_analysis}.
We carry out hyperparameter tuning for both the response pair selection methods from \cref{sec:response_pair_acquisition} and the training methods used for evaluation. Further implementation details are provided in \cref{app:implementation_details}.

\begin{figure*}[t]
    \centering
    \includegraphics{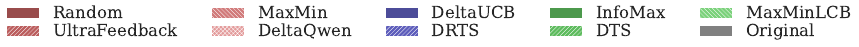}

    \begin{subfigure}[b]{0.5\textwidth}
        \centering
        \includegraphics[trim={0 0 0 0}, clip]{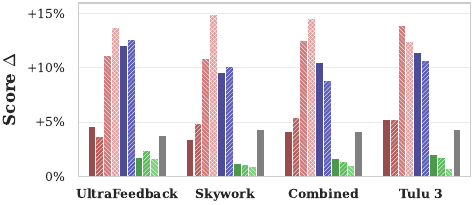}
        \caption{DPO Models}
        \label{fig:prompt_dataset_ablation_dpo}
    \end{subfigure}%
    \hfill 
    \begin{subfigure}[b]{0.5\textwidth}
        \centering
        \includegraphics[trim={0 0 0 0}, clip]{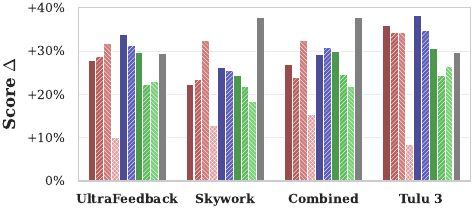}
        \caption{Reward Models}
        \label{fig:prompt_dataset_ablation_rm}
    \end{subfigure}
    
    \caption{Benchmarking of downstream and reward model performance across input prompt datasets, increasing in scale from left to right. Scores are reported as relative deltas to the base model. We provide the scores achieved using the original preference dataset instead of just the prompts with \activeuf{} for reference.} 
    \label{fig:prompt_dataset_ablation}
\end{figure*}

\subsection{Response Pair Selection Methods}
\label{sec:acquisition_evaluation}

\looseness=-1
In this section, we address our first research question by employing the \activeuf{} pipeline with the response pair selection methods described in \cref{sec:response_pair_acquisition}.
The results presented in \cref{tab:acquisition_function_comparison} and \cref{fig:teaser} demonstrate that \activeuf{} with \textsc{DRTS} and \textsc{DeltaUCB} can generate high-quality datasets for both reward modeling and preference optimization, outperforming all other methods except \textsc{DeltaQwen} for the latter.
This is expected due to the known performance of \textsc{DeltaQwen} for fine-tuning with DPO on common domains and datasets. However, it significantly lags behind even random sampling for reward modelling.
We attribute this discrepancy for \textsc{DeltaQwen} to its confinement to the training distribution of the underlying models.

Contrary to many prior works considering active learning for RLHF as a contextual dueling bandit problem (\cref{sec:related_work}), we find that previously proposed dueling bandit methods do not transfer effectively to the task of preference data generation. Analyzing the generated datasets (\cref{app:generated_dataset_analysis}) confirms that \textsc{DTS} and \textsc{MaxMinLCB} successfully achieve their theoretical goal of identifying high-quality responses, but yield datasets that lack the quality deltas required for learning.
Consequently, these methods underperform even random sampling, demonstrating that the objectives of regret minimization and uncertainty minimization are misaligned with the goal of preference data generation.
Intuitively, delta-based selection improves dataset quality by creating clearer preference boundaries. Larger quality gaps provide a lower-noise training signal than ambiguous comparisons between similarly strong answers. Appendix \cref{app:generated_dataset_analysis} supports this interpretation: \textsc{DRTS} and \textsc{DeltaUCB} retain high chosen-response scores while pairing them with substantially weaker rejected responses, whereas regret-minimizing methods such as \textsc{DTS} and \textsc{MaxMinLCB} often compare two strong answers and thus provide weaker supervision.

\begin{figure*}[ht]
    \centering
    \includegraphics{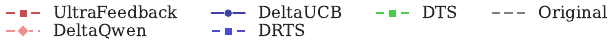}

    \begin{subfigure}[b]{0.5\textwidth}
        \centering
        \includegraphics[trim={0 0 0 0}, clip]{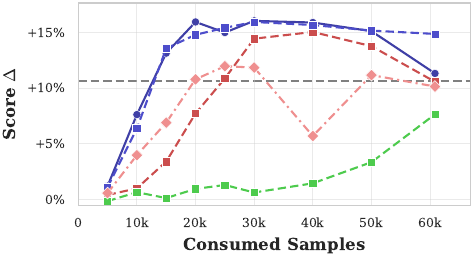}
        \caption{IPO}
        \label{fig:po_ablation_ipo_simpo_sample_effiency_ipo}
    \end{subfigure}%
    \hfill 
    \begin{subfigure}[b]{0.5\textwidth}
        \centering
        \includegraphics[trim={0 0 0 0}, clip]{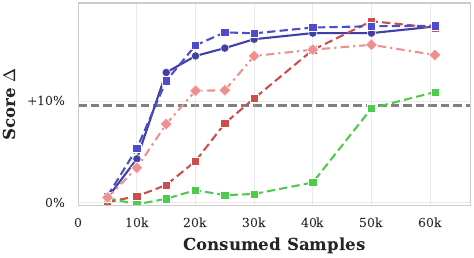}
        \caption{SimPO}
        \label{fig:po_ablation_ipo_simpo_sample_effiency_simpo}
    \end{subfigure}
    
    \caption{Mean performance trajectories for our fine-tuned models using IPO (\cref{fig:po_ablation_ipo_simpo_sample_effiency_ipo}) and SimPO (\cref{fig:po_ablation_ipo_simpo_sample_effiency_simpo}) as a function of consumed samples on datasets generated using \activeuf{} based on \textbf{UltraFeedback} prompts. For readability, we subset the selection strategies, for a full comparison, see~\cref{fig:full_po_ablation_ipo_simpo_sample_effiency} in the Appendix. We provide the scores achieved using the original preference dataset instead of just the prompts with \activeuf{} for reference.} %
    \label{fig:po_ablation_ipo_simpo_sample_effiency}
\end{figure*}

\subsection{Sample Efficiency}
\label{sec:sample_efficiency_evaluation}

We address our second research question by evaluating models trained on subsets of the generated datasets.
The results for downstream benchmarks (\cref{fig:sample_efficiency_dpo}) show that our proposed methods, \textsc{DRTS} and \textsc{DeltaUCB}, demonstrate strong sample-efficiency in downstream evaluations. Using our proposed methods, models fine-tuned on merely 5'000 to 10'000 samples outperform those trained on 60'000 samples from the datasets generated using \textsc{Random}, \textsc{UltraFeedback}, or dueling bandit methods. Notably, they also lead to better performance than when training on the original \textbf{UltraFeedback} dataset~\citep{cui2024ultrafeedbackboostinglanguagemodels}. While \textsc{DeltaQwen} shows a 1\% improvement in mean downstream score over \textsc{DRTS}, this is driven disproportionately by AlpacaEval 2 performance, as also shown on \cref{tab:acquisition_function_comparison} (see Appendix, \cref{fig:sample_efficiency_without_alpacaeval}).
Notably, \textsc{DeltaUCB} shows smaller fluctuations in performance than \textsc{MaxMin}, \textsc{DeltaQwen}, and \textsc{DRTS}.
These results indicate that DPO training can be made significantly more sample-efficient than previously reported by leveraging optimal selection of responses, and that training models on preference feedback could be achieved at a much lower annotation cost.
Our gains are primarily in \emph{annotation efficiency}, not in minimizing raw response-generation cost. In the current implementation, we pre-computed responses and judge scores for the full prompt set to support large-scale ablations, increasing upfront compute relative to static pipelines. However, strong selection methods such as \textsc{DRTS} and \textsc{DeltaUCB} achieve similar downstream performance with substantially fewer selected pairs, reducing end-to-end data collection when annotation is the main bottleneck. Raw compute estimates are given in \cref{app:compute_estimates}.

\looseness=-1
As shown on \cref{fig:sample_efficiency_rm}, reward modeling follows a more gradual saturation curve, requiring 40'000 samples to attain benchmark scores equivalent to training on the complete dataset without active response pair selection.
Furthermore, \cref{fig:sample_efficiency} reveals a critical limitation of the \textsc{DeltaQwen} baseline: its strong downstream performance (\cref{fig:sample_efficiency_dpo}) contrasts with poor generalization in reward modeling (\cref{fig:sample_efficiency_rm}).
In addition, \textsc{Random} shows strong performance for reward modeling, which, in turn, suggests that diversity is a more desirable property for this task than qualitative difference.
On the contrary, \textsc{DRTS} and \textsc{DeltaUCB} not only achieve high scores on both tasks but only these two methods are both practical and yield datasets that can surpass the quality of the original one.

\subsection{Input Prompt Dataset Ablation} \label{sec:input_prompt_dataset_ablation}

To assess the generalization capabilities of \activeuf{} beyond the \textbf{UltraFeedback} prompts, we evaluate the pipeline on three additional datasets of varying scales: 
\begin{enumerate*}[label=(\roman*)]
    \item \textbf{Skywork} Reward Preference 80k v0.2\footnote{\href{https://huggingface.co/datasets/Skywork/Skywork-Reward-Preference-80K-v0.2}{Skywork/Skywork-Reward-Preference-80K-v0.2}}~\citep{liu2024skywork}, a high-quality dataset of 80'000 prompts for reward modeling;
    \item \textbf{Combined}: a combination of the \textbf{UltraFeedback} and \textbf{Skywork} datasets with 140'000 prompts; and
    \item \textbf{Tulu 3} 8B Preference Mixture\footnote{\href{https://huggingface.co/datasets/allenai/llama-3.1-tulu-3-8b-preference-mixture}{allenai/llama-3.1-tulu-3-8b-preference-mixture}}, a dataset of 272'000 prompts for LLM fine-tuning~\citep{lambert2025tulu3pushingfrontiers}.
\end{enumerate*}

\cref{fig:prompt_dataset_ablation} confirms that \activeuf, combined with our \textsc{DRTS} and \textsc{DeltaUCB} methods, generalizes effectively across diverse prompt datasets, consistently outperforming existing preference data generation heuristics and standard methods. 
While \textsc{DeltaQwen} achieves a high downstream score, similar to \cref{sec:sample_efficiency_evaluation}, this performance is skewed by AlpacaEval 2 (see \cref{tab:prompt_dataset_ablation} for exact scores). \textsc{DeltaQwen} still significantly underperforms on RewardBench 2, which we, again, attribute to a lack of diversity.

Remarkably, our pipeline demonstrates substantial improvements over the widely-adopted original preference datasets included in \cref{fig:prompt_dataset_ablation} (\textbf{UltraFeedback}, \textbf{Skywork}, and \textbf{Tulu 3}). In terms of DPO mean scores, \textsc{DRTS} and \textsc{DeltaUCB} yield significantly better results across all prompt sources. While the reference \textbf{Skywork} and \textbf{Combined} datasets retain an advantage in reward model training, which is expected as \textbf{Skywork} is curated for reward modelling, our active delta learning methods outperform the baselines on the \textbf{UltraFeedback} and \textbf{Tulu 3} prompts.

\subsection{Preference Optimization Algorithm Ablation} \label{sec:preference_optimization_algorithm_ablation}

\looseness=-1
To evaluate the generalizability of \activeuf{} across different preference optimization algorithms beyond DPO~\citep{rafailov2023direct}, we extend our analysis in \cref{sec:acquisition_evaluation} to the IPO~\citep{du2024ipo} and SimPO~\citep{meng2024simpo} algorithms. While DPO optimizes the policy by implicitly maximizing a reward function with KL-regularization, IPO maximizes the win rate against a fixed policy, eliminating the need for a reward model, and SimPO simplifies the objective by using a length-normalized reward margin for regularization. The results are visualized in \cref{fig:po_ablation_ipo_simpo_sample_effiency}.

Regardless of the optimization algorithm, \textsc{DRTS} and \textsc{DeltaUCB} remain among the highest performing methods, and their trajectories demonstrate the superior sample efficiency by converging to their top performance using significantly fewer samples than all other methods.
In contrast, \textsc{DeltaQwen} suffers a significant performance drop on these alternative algorithms, demonstrating its inflexibility and limiting its applicability to very specific experimental setups.
We attribute this behavior to the limited diversity of the \textsc{DeltaQwen} datasets. Unlike \textsc{DRTS} and \textsc{DeltaUCB}, which select from the full model pool, \textsc{DeltaQwen} always compares the same two Qwen models. This seems sufficient for DPO on common domains, but less robust for IPO and SimPO, which are more sensitive to drift and over-specialization. This is consistent with our diversity analysis in \cref{app:generated_dataset_analysis}.

We observe that \textsc{Random}, \textsc{UltraFeedback}, and \textsc{DTS} perform remarkably well with IPO and SimPO, compared to their performance with DPO, but they achieve high performance with large datasets only.
Detailed numerical results are provided in \cref{app:full_preference_optimization_algorithm_ablation} and \cref{tab:preference_optimization_algorithm_ablation}.

\section{Conclusion}

We present \activeuf, a modular active learning pipeline for preference data generation. \activeuf{} addresses a central bottleneck in preference optimization: selecting the most informative response pairs for labeling within a limited annotation budget.
Our extensive evaluations demonstrate that using datasets produced by \activeuf, particularly when coupled with our novel \textsc{DRTS} and \textsc{DeltaUCB} response selection methods, results in significantly stronger reward and fine-tuned models compared to those derived from static heuristics. Notably, these gains are consistent across varying prompt sources and optimization algorithms, making our approach the first to produce high-quality datasets agnostic to the downstream task or training algorithm. In practical terms, our results suggest three takeaways. First, when annotation budget is limited, \textsc{DRTS} and \textsc{DeltaUCB} are strong default choices. Second, selecting the single best response is not the same as selecting the most informative response pair. Third, while \textsc{DeltaQwen} is competitive for DPO on common domains, it generalizes less reliably across tasks. Overall, we recommend \textsc{ActiveUltraFeedback} with \textsc{DRTS} or \textsc{DeltaUCB} for efficient preference-data collection.

Importantly, \activeuf{} is designed as a \emph{platform} for preference-data collection, enabling researchers and practitioners to rapidly develop, swap, and benchmark new methods, uncertainty estimators, and judges.
We see many promising directions of future work to build on this platform, such as testing additional uncertainty estimation approaches, setting explicit diversity constraints, incorporating prompt selection into the active learning loop, creating open-source datasets for expert and low-resource domains, and extending the platform with a user interface to collect human annotations. Furthermore, we recognize that the current pipeline incurs substantial computational cost due to generating responses from many LLMs for each prompt. Therefore, we see strong potential in selecting models to query for responses instead of selecting between already generated responses as a high priority. To lower the barrier to entry and make this line of research more accessible, we therefore release all generated datasets, enabling future researchers to build upon our results without incurring the full computational overhead.

\section*{Acknowledgments}

This work was supported as part of the Swiss AI initiative by a grant from the Swiss National Supercomputing Centre (CSCS) on Alps.
Barna P\'{a}sztor was primarily supported by the ETH AI Center through an ETH AI Center doctoral fellowship, and Ido Hakimi was primarily supported by the ETH AI Center through an ETH AI Center postdoctoral fellowship.

\section*{Impact Statement}
This paper presents \activeuf, an active learning pipeline for preference-data collection in RLHF that improves sample efficiency and reduces reliance on human annotation, potentially broadening access to preference optimization and enabling faster iteration on alignment datasets across diverse domains. As with other preference-based approaches, \activeuf{} may amplify biases in prompts, annotators, or judges, and stronger reward models may increase the risk of reward hacking or over-optimization; while it does not introduce new capabilities for generating harmful content, it could be misused to more efficiently optimize models toward undesirable preferences. We mitigate these risks through evaluation across diverse prompt sources and benchmarks, release of code and datasets for reproducibility and auditing, and a modular design that allows practitioners to incorporate improved judges, safety filters, and bias-mitigation strategies. We encourage future deployments to pair preference-data collection with clear annotation guidelines, safety-focused evaluations, and monitoring for distribution shift and reward-model failures.

\newpage
\bibliography{references}

@InProceedings{cui2024ultrafeedbackboostinglanguagemodels,
  title = 	 {{ULTRAFEEDBACK}: Boosting Language Models with Scaled {AI} Feedback},
  author =       {Cui, Ganqu and Yuan, Lifan and Ding, Ning and Yao, Guanming and He, Bingxiang and Zhu, Wei and Ni, Yuan and Xie, Guotong and Xie, Ruobing and Lin, Yankai and Liu, Zhiyuan and Sun, Maosong},
  booktitle = 	 {Proceedings of the 41st International Conference on Machine Learning},
  pages = 	 {9722--9744},
  year = 	 {2024},
  editor = 	 {Salakhutdinov, Ruslan and Kolter, Zico and Heller, Katherine and Weller, Adrian and Oliver, Nuria and Scarlett, Jonathan and Berkenkamp, Felix},
  volume = 	 {235},
  series = 	 {Proceedings of Machine Learning Research},
  month = 	 {21--27 Jul},
  publisher =    {PMLR},
  url = 	 {https://proceedings.mlr.press/v235/cui24f.html}
}

@article{gemmateam2024gemmaopenmodelsbased,
  title={Gemma: Open Models Based on Gemini Research and Technology},
  author={Team, Gemma and Mesnard, Thomas and Hardin, Cassidy and Dadashi, Robert and Bhupatiraju, Surya and Pathak, Shreya and Sifre, Laurent and Rivi{\`e}re, Morgane and Kale, Mihir Sanjay and Love, Juliette and others},
  journal={arXiv preprint arXiv:2403.08295},
  year={2024}
}

@inproceedings{ouyang2022training,
 author = {Ouyang, Long and Wu, Jeffrey and Jiang, Xu and Almeida, Diogo and Wainwright, Carroll and Mishkin, Pamela and Zhang, Chong and Agarwal, Sandhini and Slama, Katarina and Ray, Alex and Schulman, John and Hilton, Jacob and Kelton, Fraser and Miller, Luke and Simens, Maddie and Askell, Amanda and Welinder, Peter and Christiano, Paul F and Leike, Jan and Lowe, Ryan},
 booktitle = {Advances in Neural Information Processing Systems},
 editor = {S. Koyejo and S. Mohamed and A. Agarwal and D. Belgrave and K. Cho and A. Oh},
 pages = {27730--27744},
 publisher = {Curran Associates, Inc.},
 title = {Training language models to follow instructions with human feedback},
 url = {https://proceedings.neurips.cc/paper_files/paper/2022/file/b1efde53be364a73914f58805a001731-Paper-Conference.pdf},
 volume = {35},
 year = {2022}
}

@article{bradley1952rank,
  title={Rank analysis of incomplete block designs: I. the method of paired comparisons},
  author={Bradley, Ralph Allan and Terry, Milton E},
  journal={Biometrika},
  volume={39},
  number={3/4},
  pages={324--345},
  year={1952},
  publisher={JSTOR}
}

@inproceedings{
tunstall2024zephyr,
title={Zephyr: Direct Distillation of {LM} Alignment},
author={Lewis Tunstall and Edward Emanuel Beeching and Nathan Lambert and Nazneen Rajani and Kashif Rasul and Younes Belkada and Shengyi Huang and Leandro Von Werra and Cl{\'e}mentine Fourrier and Nathan Habib and Nathan Sarrazin and Omar Sanseviero and Alexander M Rush and Thomas Wolf},
booktitle={First Conference on Language Modeling},
year={2024},
url={https://openreview.net/forum?id=aKkAwZB6JV}
}

@inproceedings{wang2024helpsteer,
 author = {Wang, Zhilin and Dong, Yi and Delalleau, Olivier and Zeng, Jiaqi and Shen, Gerald and Egert, Daniel and Zhang, Jimmy J. and Sreedhar, Makesh Narsimhan and Kuchaiev, Oleksii},
 booktitle = {Advances in Neural Information Processing Systems},
 doi = {10.52202/079017-0047},
 editor = {A. Globerson and L. Mackey and D. Belgrave and A. Fan and U. Paquet and J. Tomczak and C. Zhang},
 pages = {1474--1501},
 publisher = {Curran Associates, Inc.},
 title = {HelpSteer 2: Open-source dataset for training top-performing reward models},
 url = {https://proceedings.neurips.cc/paper_files/paper/2024/file/02fd91a387a6a5a5751e81b58a75af90-Paper-Datasets_and_Benchmarks_Track.pdf},
 volume = {37},
 year = {2024}
}

@InProceedings{pmlr-v162-ethayarajh22a,
  title = 	 {Understanding Dataset Difficulty with $\mathcal{V}$-Usable Information},
  author =       {Ethayarajh, Kawin and Choi, Yejin and Swayamdipta, Swabha},
  booktitle = 	 {Proceedings of the 39th International Conference on Machine Learning},
  pages = 	 {5988--6008},
  year = 	 {2022},
  editor = 	 {Chaudhuri, Kamalika and Jegelka, Stefanie and Song, Le and Szepesvari, Csaba and Niu, Gang and Sabato, Sivan},
  volume = 	 {162},
  series = 	 {Proceedings of Machine Learning Research},
  month = 	 {17--23 Jul},
  publisher =    {PMLR},
  url = 	 {https://proceedings.mlr.press/v162/ethayarajh22a.html}
}

@inproceedings{
    starling2023,
    title={Starling-7B: Improving Helpfulness and Harmlessness with {RLAIF}},
    author={Banghua Zhu and Evan Frick and Tianhao Wu and Hanlin Zhu and Karthik Ganesan and Wei-Lin Chiang and Jian Zhang and Jiantao Jiao},
    booktitle={First Conference on Language Modeling},
    year={2024},
    url={https://openreview.net/forum?id=GqDntYTTbk}
}

@inproceedings{wang2024interpretable,
    title = "Interpretable Preferences via Multi-Objective Reward Modeling and Mixture-of-Experts",
    author = "Wang, Haoxiang  and
      Xiong, Wei  and
      Xie, Tengyang  and
      Zhao, Han  and
      Zhang, Tong",
    editor = "Al-Onaizan, Yaser  and
      Bansal, Mohit  and
      Chen, Yun-Nung",
    booktitle = "Findings of the Association for Computational Linguistics: EMNLP 2024",
    month = nov,
    year = "2024",
    address = "Miami, Florida, USA",
    publisher = "Association for Computational Linguistics",
    url = "https://aclanthology.org/2024.findings-emnlp.620/",
    doi = "10.18653/v1/2024.findings-emnlp.620",
    pages = "10582--10592"
}

@inproceedings{osband2023epistemicneuralnetworks,
 author = {Osband, Ian and Wen, Zheng and Asghari, Seyed Mohammad and Dwaracherla, Vikranth and IBRAHIMI, MORTEZA and Lu, Xiuyuan and Van Roy, Benjamin},
 booktitle = {Advances in Neural Information Processing Systems},
 editor = {A. Oh and T. Naumann and A. Globerson and K. Saenko and M. Hardt and S. Levine},
 pages = {2795--2823},
 publisher = {Curran Associates, Inc.},
 title = {Epistemic Neural Networks},
 url = {https://proceedings.neurips.cc/paper_files/paper/2023/file/07fbde96bee50f4e09303fd4f877c2f3-Paper-Conference.pdf},
 volume = {36},
 year = {2023}
}

@inproceedings{
lambert2025tulu3pushingfrontiers,
title={Tulu 3: Pushing Frontiers in Open Language Model Post-Training},
author={Nathan Lambert and Jacob Morrison and Valentina Pyatkin and Shengyi Huang and Hamish Ivison and Faeze Brahman and Lester James Validad Miranda and Alisa Liu and Nouha Dziri and Xinxi Lyu and Yuling Gu and Saumya Malik and Victoria Graf and Jena D. Hwang and Jiangjiang Yang and Ronan Le Bras and Oyvind Tafjord and Christopher Wilhelm and Luca Soldaini and Noah A. Smith and Yizhong Wang and Pradeep Dasigi and Hannaneh Hajishirzi},
booktitle={Second Conference on Language Modeling},
year={2025},
url={https://openreview.net/forum?id=i1uGbfHHpH}
}

@inproceedings{wu2016doublethompsonsamplingdueling,
 author = {Wu, Huasen and Liu, Xin},
 booktitle = {Advances in Neural Information Processing Systems},
 editor = {D. Lee and M. Sugiyama and U. Luxburg and I. Guyon and R. Garnett},
 pages = {},
 publisher = {Curran Associates, Inc.},
 title = {Double Thompson Sampling for Dueling Bandits},
 url = {https://proceedings.neurips.cc/paper_files/paper/2016/file/9de6d14fff9806d4bcd1ef555be766cd-Paper.pdf},
 volume = {29},
 year = {2016}
}

@InProceedings{dwaracherla2024efficientexplorationllms,
  title = 	 {Efficient Exploration for {LLM}s},
  author =       {Dwaracherla, Vikranth and Asghari, Seyed Mohammad and Hao, Botao and Van Roy, Benjamin},
  booktitle = 	 {Proceedings of the 41st International Conference on Machine Learning},
  pages = 	 {12215--12227},
  year = 	 {2024},
  editor = 	 {Salakhutdinov, Ruslan and Kolter, Zico and Heller, Katherine and Weller, Adrian and Oliver, Nuria and Scarlett, Jonathan and Berkenkamp, Felix},
  volume = 	 {235},
  series = 	 {Proceedings of Machine Learning Research},
  month = 	 {21--27 Jul},
  publisher =    {PMLR},
  url = 	 {https://proceedings.mlr.press/v235/dwaracherla24a.html}
}

@inproceedings{melo2024deep,
 author = {Melo, Luckeciano C. and Tigas, Panagiotis and Abate, Alessandro and Gal, Yarin},
 booktitle = {Advances in Neural Information Processing Systems},
 doi = {10.52202/079017-3749},
 editor = {A. Globerson and L. Mackey and D. Belgrave and A. Fan and U. Paquet and J. Tomczak and C. Zhang},
 pages = {118052--118085},
 publisher = {Curran Associates, Inc.},
 title = {Deep Bayesian Active Learning for Preference Modeling in Large Language Models},
 url = {https://proceedings.neurips.cc/paper_files/paper/2024/file/d5e256c988bdee59a0f4d7a9bc1dd6d9-Paper-Conference.pdf},
 volume = {37},
 year = {2024}
}

@inproceedings{
liu2024sampleefficientalignmentllms,
title={Sample-Efficient Alignment for {LLM}s},
author={Zichen Liu and Changyu Chen and Chao Du and Wee Sun Lee and Min Lin},
booktitle={Language Gamification - NeurIPS 2024 Workshop},
year={2024},
url={https://openreview.net/forum?id=6Kcvz310CX}
}

@inproceedings{pasztor2024bandits,
 author = {P\'{a}sztor, Barna and Kassraie, Parnian and Krause, Andreas},
 booktitle = {Advances in Neural Information Processing Systems},
 doi = {10.52202/079017-0383},
 editor = {A. Globerson and L. Mackey and D. Belgrave and A. Fan and U. Paquet and J. Tomczak and C. Zhang},
 pages = {11997--12034},
 publisher = {Curran Associates, Inc.},
 title = {Bandits with Preference Feedback: A Stackelberg Game Perspective},
 url = {https://proceedings.neurips.cc/paper_files/paper/2024/file/1646e34971facbcda3727d1dc28ab635-Paper-Conference.pdf},
 volume = {37},
 year = {2024}
}

@misc{malik2025rewardbench2advancingreward,
      title={RewardBench 2: Advancing Reward Model Evaluation}, 
      author={Saumya Malik and Valentina Pyatkin and Sander Land and Jacob Morrison and Noah A. Smith and Hannaneh Hajishirzi and Nathan Lambert},
      year={2025},
      eprint={2506.01937},
      archivePrefix={arXiv},
      primaryClass={cs.CL},
      url={https://arxiv.org/abs/2506.01937}, 
}

@inproceedings{
olmo20252olmo2furious,
title={2 {OLM}o 2 Furious ({COLM}{\textquoteright}s Version)},
author={Evan Pete Walsh and Luca Soldaini and Dirk Groeneveld and Kyle Lo and Shane Arora and Akshita Bhagia and Yuling Gu and Shengyi Huang and Matt Jordan and Nathan Lambert and Dustin Schwenk and Oyvind Tafjord and Taira Anderson and David Atkinson and Faeze Brahman and Christopher Clark and Pradeep Dasigi and Nouha Dziri and Allyson Ettinger and Michal Guerquin and David Heineman and Hamish Ivison and Pang Wei Koh and Jiacheng Liu and Saumya Malik and William Merrill and Lester James Validad Miranda and Jacob Morrison and Tyler Murray and Crystal Nam and Jake Poznanski and Valentina Pyatkin and Aman Rangapur and Michael Schmitz and Sam Skjonsberg and David Wadden and Christopher Wilhelm and Michael Wilson and Luke Zettlemoyer and Ali Farhadi and Noah A. Smith and Hannaneh Hajishirzi},
booktitle={Second Conference on Language Modeling},
year={2025},
url={https://openreview.net/forum?id=2ezugTT9kU}
}

@article{grattafiori2024llama3herdmodels,
  title={The llama 3 herd of models},
  author={Grattafiori, Aaron and Dubey, Abhimanyu and Jauhri, Abhinav and Pandey, Abhinav and Kadian, Abhishek and Al-Dahle, Ahmad and Letman, Aiesha and Mathur, Akhil and Schelten, Alan and Vaughan, Alex and others},
  journal={arXiv preprint arXiv:2407.21783},
  year={2024}
}

@misc{zhou2023instructionfollowingevaluationlargelanguage,
      title={Instruction-Following Evaluation for Large Language Models}, 
      author={Jeffrey Zhou and Tianjian Lu and Swaroop Mishra and Siddhartha Brahma and Sujoy Basu and Yi Luan and Denny Zhou and Le Hou},
      year={2023},
      eprint={2311.07911},
      archivePrefix={arXiv},
      primaryClass={cs.CL},
      url={https://arxiv.org/abs/2311.07911}, 
}

@inproceedings{rafailov2023direct,
 author = {Rafailov, Rafael and Sharma, Archit and Mitchell, Eric and Manning, Christopher D and Ermon, Stefano and Finn, Chelsea},
 booktitle = {Advances in Neural Information Processing Systems},
 editor = {A. Oh and T. Naumann and A. Globerson and K. Saenko and M. Hardt and S. Levine},
 pages = {53728--53741},
 publisher = {Curran Associates, Inc.},
 title = {Direct Preference Optimization: Your Language Model is Secretly a Reward Model},
 url = {https://proceedings.neurips.cc/paper_files/paper/2023/file/a85b405ed65c6477a4fe8302b5e06ce7-Paper-Conference.pdf},
 volume = {36},
 year = {2023}
}

@inproceedings{dudik2011efficientoptimallearningcontextual,
author = {Dudik, Miroslav and Hsu, Daniel and Kale, Satyen and Karampatziakis, Nikos and Langford, John and Reyzin, Lev and Zhang, Tong},
title = {Efficient optimal learning for contextual bandits},
year = {2011},
isbn = {9780974903972},
publisher = {AUAI Press},
address = {Arlington, Virginia, USA},
booktitle = {Proceedings of the Twenty-Seventh Conference on Uncertainty in Artificial Intelligence},
pages = {169–178},
numpages = {10},
location = {Barcelona, Spain},
series = {UAI'11}
}

@InProceedings{dudik2015contextualduelingbandits,
  title = 	 {Contextual Dueling Bandits},
  author = 	 {Dud{\'\i}k, Miroslav and Hofmann, Katja and Schapire, Robert E. and Slivkins, Aleksandrs and Zoghi, Masrour},
  booktitle = 	 {Proceedings of The 28th Conference on Learning Theory},
  pages = 	 {563--587},
  year = 	 {2015},
  editor = 	 {Grünwald, Peter and Hazan, Elad and Kale, Satyen},
  volume = 	 {40},
  series = 	 {Proceedings of Machine Learning Research},
  address = 	 {Paris, France},
  month = 	 {03--06 Jul},
  publisher =    {PMLR},
  url = 	 {https://proceedings.mlr.press/v40/Dudik15.html}
}

@article{bengs2021preference,
  author  = {Viktor Bengs and R{\'o}bert Busa-Fekete and Adil El Mesaoudi-Paul and Eyke H{\"u}llermeier},
  title   = {Preference-based Online Learning with Dueling Bandits: A Survey},
  journal = {Journal of Machine Learning Research},
  year    = {2021},
  volume  = {22},
  number  = {7},
  pages   = {1--108},
  url     = {http://jmlr.org/papers/v22/18-546.html}
}

@article{liu2024skywork,
  title={Skywork-Reward: Bag of Tricks for Reward Modeling in LLMs},
  author={Liu, Chris Yuhao and Zeng, Liang and Liu, Jiacai and Yan, Rui and He, Jujie and Wang, Chaojie and Yan, Shuicheng and Liu, Yang and Zhou, Yahui},
  journal={arXiv preprint arXiv:2410.18451},
  year={2024}
}

@InProceedings{muldrew2024activepreferencelearninglarge,
  title = 	 {Active Preference Learning for Large Language Models},
  author =       {Muldrew, William and Hayes, Peter and Zhang, Mingtian and Barber, David},
  booktitle = 	 {Proceedings of the 41st International Conference on Machine Learning},
  pages = 	 {36577--36590},
  year = 	 {2024},
  editor = 	 {Salakhutdinov, Ruslan and Kolter, Zico and Heller, Katherine and Weller, Adrian and Oliver, Nuria and Scarlett, Jonathan and Berkenkamp, Felix},
  volume = 	 {235},
  series = 	 {Proceedings of Machine Learning Research},
  month = 	 {21--27 Jul},
  publisher =    {PMLR},
  url = 	 {https://proceedings.mlr.press/v235/muldrew24a.html}
}

@article{zhai2023uncertaintypenalizedreinforcementlearninghuman,
title = {Uncertainty-penalized reinforcement learning from human feedback with diversified reward LoRA ensembles},
journal = {Information Processing \& Management},
volume = {63},
number = {3},
pages = {104548},
year = {2026},
issn = {0306-4573},
doi = {https://doi.org/10.1016/j.ipm.2025.104548},
url = {https://www.sciencedirect.com/science/article/pii/S0306457325004893},
author = {Yuanzhao Zhai and Yu Lei and Han Zhang and Yue Yu and Kele Xu and Dawei Feng and Bo Ding and Huaimin Wang}
}

@inproceedings{geng2025deltalearninghypothesispreference,
title={The Delta Learning Hypothesis: Preference Tuning on Weak Data can Yield Strong Gains},
author={Scott Geng and Hamish Ivison and Chun-Liang Li and Maarten Sap and Jerry Li and Ranjay Krishna and Pang Wei Koh},
booktitle={Second Conference on Language Modeling},
year={2025},
url={https://openreview.net/forum?id=9rwtezthwo}
}

@article{shen2025reviving,
  title={Reviving the classics: Active reward modeling in large language model alignment},
  author={Shen, Yunyi and Sun, Hao and Ton, Jean-Fran{\c{c}}ois},
  journal={arXiv preprint arXiv:2502.04354},
  year={2025}
}

@inproceedings{christiano2017deep,
 author = {Christiano, Paul F and Leike, Jan and Brown, Tom and Martic, Miljan and Legg, Shane and Amodei, Dario},
 booktitle = {Advances in Neural Information Processing Systems},
 editor = {I. Guyon and U. Von Luxburg and S. Bengio and H. Wallach and R. Fergus and S. Vishwanathan and R. Garnett},
 pages = {},
 publisher = {Curran Associates, Inc.},
 title = {Deep Reinforcement Learning from Human Preferences},
 url = {https://proceedings.neurips.cc/paper_files/paper/2017/file/d5e2c0adad503c91f91df240d0cd4e49-Paper.pdf},
 volume = {30},
 year = {2017}
}

@article{schulman2017proximal,
  title={Proximal policy optimization algorithms},
  author={Schulman, John and Wolski, Filip and Dhariwal, Prafulla and Radford, Alec and Klimov, Oleg},
  journal={arXiv preprint arXiv:1707.06347},
  year={2017}
}

@article{cobbe2021training,
  title={Training verifiers to solve math word problems},
  author={Cobbe, Karl and Kosaraju, Vineet and Bavarian, Mohammad and Chen, Mark and Jun, Heewoo and Kaiser, Lukasz and Plappert, Matthias and Tworek, Jerry and Hilton, Jacob and Nakano, Reiichiro and others},
  journal={arXiv preprint arXiv:2110.14168},
  year={2021}
}

@inproceedings{lin2022truthfulqa,
    title = "{T}ruthful{QA}: Measuring How Models Mimic Human Falsehoods",
    author = "Lin, Stephanie  and
      Hilton, Jacob  and
      Evans, Owain",
    editor = "Muresan, Smaranda  and
      Nakov, Preslav  and
      Villavicencio, Aline",
    booktitle = "Proceedings of the 60th Annual Meeting of the Association for Computational Linguistics (Volume 1: Long Papers)",
    month = may,
    year = "2022",
    address = "Dublin, Ireland",
    publisher = "Association for Computational Linguistics",
    url = "https://aclanthology.org/2022.acl-long.229/",
    doi = "10.18653/v1/2022.acl-long.229",
    pages = "3214--3252"
}

@inproceedings{
hu2022lora,
title={Lo{RA}: Low-Rank Adaptation of Large Language Models},
author={Edward J Hu and yelong shen and Phillip Wallis and Zeyuan Allen-Zhu and Yuanzhi Li and Shean Wang and Lu Wang and Weizhu Chen},
booktitle={International Conference on Learning Representations},
year={2022},
url={https://openreview.net/forum?id=nZeVKeeFYf9}
}

@article{dubois2024length,
  title={Length-Controlled AlpacaEval: A Simple Way to Debias Automatic Evaluators},
  author={Dubois, Yann and Galambosi, Bal{\'a}zs and Liang, Percy and Hashimoto, Tatsunori B},
  journal={arXiv preprint arXiv:2404.04475},
  year={2024}
}

@inproceedings{meng2024simpo,
 author = {Meng, Yu and Xia, Mengzhou and Chen, Danqi},
 booktitle = {Advances in Neural Information Processing Systems},
 doi = {10.52202/079017-3946},
 editor = {A. Globerson and L. Mackey and D. Belgrave and A. Fan and U. Paquet and J. Tomczak and C. Zhang},
 pages = {124198--124235},
 publisher = {Curran Associates, Inc.},
 title = {SimPO: Simple Preference Optimization with a Reference-Free Reward},
 url = {https://proceedings.neurips.cc/paper_files/paper/2024/file/e099c1c9699814af0be873a175361713-Paper-Conference.pdf},
 volume = {37},
 year = {2024}
}

@inproceedings{du2024ipo,
 author = {Du, Yingjun and Sun, Wenfang and Snoek, Cees G. M.},
 booktitle = {Advances in Neural Information Processing Systems},
 doi = {10.52202/079017-4025},
 editor = {A. Globerson and L. Mackey and D. Belgrave and A. Fan and U. Paquet and J. Tomczak and C. Zhang},
 pages = {126725--126766},
 publisher = {Curran Associates, Inc.},
 title = {IPO: Interpretable Prompt Optimization for Vision-Language Models},
 url = {https://proceedings.neurips.cc/paper_files/paper/2024/file/e52e4de8689a9955b6d3ff421d019387-Paper-Conference.pdf},
 volume = {37},
 year = {2024}
}

@misc{olmo2025olmo,
      title={Olmo 3}, 
      author={Team Olmo and : and Allyson Ettinger and Amanda Bertsch and Bailey Kuehl and David Graham and David Heineman and Dirk Groeneveld and Faeze Brahman and Finbarr Timbers and Hamish Ivison and Jacob Morrison and Jake Poznanski and Kyle Lo and Luca Soldaini and Matt Jordan and Mayee Chen and Michael Noukhovitch and Nathan Lambert and Pete Walsh and Pradeep Dasigi and Robert Berry and Saumya Malik and Saurabh Shah and Scott Geng and Shane Arora and Shashank Gupta and Taira Anderson and Teng Xiao and Tyler Murray and Tyler Romero and Victoria Graf and Akari Asai and Akshita Bhagia and Alexander Wettig and Alisa Liu and Aman Rangapur and Chloe Anastasiades and Costa Huang and Dustin Schwenk and Harsh Trivedi and Ian Magnusson and Jaron Lochner and Jiacheng Liu and Lester James V. Miranda and Maarten Sap and Malia Morgan and Michael Schmitz and Michal Guerquin and Michael Wilson and Regan Huff and Ronan Le Bras and Rui Xin and Rulin Shao and Sam Skjonsberg and Shannon Zejiang Shen and Shuyue Stella Li and Tucker Wilde and Valentina Pyatkin and Will Merrill and Yapei Chang and Yuling Gu and Zhiyuan Zeng and Ashish Sabharwal and Luke Zettlemoyer and Pang Wei Koh and Ali Farhadi and Noah A. Smith and Hannaneh Hajishirzi},
      year={2025},
      eprint={2512.13961},
      archivePrefix={arXiv},
      primaryClass={cs.CL},
      url={https://arxiv.org/abs/2512.13961}, 
}

@article{yang2025qwen3technicalreport,
  title={Qwen3 technical report},
  author={Yang, An and Li, Anfeng and Yang, Baosong and Zhang, Beichen and Hui, Binyuan and Zheng, Bo and Yu, Bowen and Gao, Chang and Huang, Chengen and Lv, Chenxu and others},
  journal={arXiv preprint arXiv:2505.09388},
  year={2025}
}

@misc{allal2025smollm2smolgoesbig,
      title={SmolLM2: When Smol Goes Big -- Data-Centric Training of a Small Language Model}, 
      author={Loubna Ben Allal and Anton Lozhkov and Elie Bakouch and Gabriel Martín Blázquez and Guilherme Penedo and Lewis Tunstall and Andrés Marafioti and Hynek Kydlíček and Agustín Piqueres Lajarín and Vaibhav Srivastav and Joshua Lochner and Caleb Fahlgren and Xuan-Son Nguyen and Clémentine Fourrier and Ben Burtenshaw and Hugo Larcher and Haojun Zhao and Cyril Zakka and Mathieu Morlon and Colin Raffel and Leandro von Werra and Thomas Wolf},
      year={2025},
      eprint={2502.02737},
      archivePrefix={arXiv},
      primaryClass={cs.CL},
      url={https://arxiv.org/abs/2502.02737}, 
}

@misc{bakouch2025smollm3,
  title={{SmolLM3: smol, multilingual, long-context reasoner}},
  author={Bakouch, Elie and Ben Allal, Loubna and Lozhkov, Anton and Tazi, Nouamane and Tunstall, Lewis and Patiño, Carlos Miguel and Beeching, Edward and Roucher, Aymeric and Reedi, Aksel Joonas and Gallouédec, Quentin and Rasul, Kashif and Habib, Nathan and Fourrier, Clémentine and Kydlicek, Hynek and Penedo, Guilherme and Larcher, Hugo and Morlon, Mathieu and Srivastav, Vaibhav and Lochner, Joshua and Nguyen, Xuan-Son and Raffel, Colin and von Werra, Leandro and Wolf, Thomas},
  year={2025},
  howpublished={\url{https://huggingface.co/blog/smollm3}}
}

@misc{kveton2025activelearningdirectpreference,
      title={Active Learning for Direct Preference Optimization}, 
      author={Branislav Kveton and Xintong Li and Julian McAuley and Ryan Rossi and Jingbo Shang and Junda Wu and Tong Yu},
      year={2025},
      eprint={2503.01076},
      archivePrefix={arXiv},
      primaryClass={cs.LG},
      url={https://arxiv.org/abs/2503.01076}, 
}

@misc{bai2022traininghelpfulharmlessassistant,
      title={Training a Helpful and Harmless Assistant with Reinforcement Learning from Human Feedback}, 
      author={Yuntao Bai and Andy Jones and Kamal Ndousse and Amanda Askell and Anna Chen and Nova DasSarma and Dawn Drain and Stanislav Fort and Deep Ganguli and Tom Henighan and Nicholas Joseph and Saurav Kadavath and Jackson Kernion and Tom Conerly and Sheer El-Showk and Nelson Elhage and Zac Hatfield-Dodds and Danny Hernandez and Tristan Hume and Scott Johnston and Shauna Kravec and Liane Lovitt and Neel Nanda and Catherine Olsson and Dario Amodei and Tom Brown and Jack Clark and Sam McCandlish and Chris Olah and Ben Mann and Jared Kaplan},
      year={2022},
      eprint={2204.05862},
      archivePrefix={arXiv},
      primaryClass={cs.CL},
      url={https://arxiv.org/abs/2204.05862}, 
}

@inproceedings{stiennon2022learningsummarizehumanfeedback,
 author = {Stiennon, Nisan and Ouyang, Long and Wu, Jeffrey and Ziegler, Daniel and Lowe, Ryan and Voss, Chelsea and Radford, Alec and Amodei, Dario and Christiano, Paul F},
 booktitle = {Advances in Neural Information Processing Systems},
 editor = {H. Larochelle and M. Ranzato and R. Hadsell and M.F. Balcan and H. Lin},
 pages = {3008--3021},
 publisher = {Curran Associates, Inc.},
 title = {Learning to summarize with human feedback},
 url = {https://proceedings.neurips.cc/paper_files/paper/2020/file/1f89885d556929e98d3ef9b86448f951-Paper.pdf},
 volume = {33},
 year = {2020}
}

@article{ziegler2019fine,
  title={Fine-tuning language models from human preferences},
  author={Ziegler, Daniel M and Stiennon, Nisan and Wu, Jeffrey and Brown, Tom B and Radford, Alec and Amodei, Dario and Christiano, Paul and Irving, Geoffrey},
  journal={arXiv preprint arXiv:1909.08593},
  year={2019}
}

@misc{qwen2025qwen25technicalreport,
      title={Qwen2.5 Technical Report}, 
      author={Qwen and : and An Yang and Baosong Yang and Beichen Zhang and Binyuan Hui and Bo Zheng and Bowen Yu and Chengyuan Li and Dayiheng Liu and Fei Huang and Haoran Wei and Huan Lin and Jian Yang and Jianhong Tu and Jianwei Zhang and Jianxin Yang and Jiaxi Yang and Jingren Zhou and Junyang Lin and Kai Dang and Keming Lu and Keqin Bao and Kexin Yang and Le Yu and Mei Li and Mingfeng Xue and Pei Zhang and Qin Zhu and Rui Men and Runji Lin and Tianhao Li and Tianyi Tang and Tingyu Xia and Xingzhang Ren and Xuancheng Ren and Yang Fan and Yang Su and Yichang Zhang and Yu Wan and Yuqiong Liu and Zeyu Cui and Zhenru Zhang and Zihan Qiu},
      year={2025},
      eprint={2412.15115},
      archivePrefix={arXiv},
      primaryClass={cs.CL},
      url={https://arxiv.org/abs/2412.15115}, 
}

@inproceedings{infobasedactiveselection,
 author = {Saha, Aadirupa},
 booktitle = {Advances in Neural Information Processing Systems},
 editor = {M. Ranzato and A. Beygelzimer and Y. Dauphin and P.S. Liang and J. Wortman Vaughan},
 pages = {30050--30062},
 publisher = {Curran Associates, Inc.},
 title = {Optimal Algorithms for Stochastic Contextual Preference Bandits },
 url = {https://proceedings.neurips.cc/paper_files/paper/2021/file/fc3cf452d3da8402bebb765225ce8c0e-Paper.pdf},
 volume = {34},
 year = {2021}
}

@inproceedings{xu2024magpiealignmentdatasynthesis,
title={Magpie: Alignment Data Synthesis from Scratch by Prompting Aligned {LLM}s with Nothing},
author={Zhangchen Xu and Fengqing Jiang and Luyao Niu and Yuntian Deng and Radha Poovendran and Yejin Choi and Bill Yuchen Lin},
booktitle={The Thirteenth International Conference on Learning Representations},
year={2025},
url={https://openreview.net/forum?id=Pnk7vMbznK}
}

@inproceedings{ivison2024unpacking,
 author = {Ivison, Hamish and Wang, Yizhong and Liu, Jiacheng and Wu, Zeqiu and Pyatkin, Valentina and Lambert, Nathan and Smith, Noah A. and Choi, Yejin and Hajishirzi, Hannaneh},
 booktitle = {Advances in Neural Information Processing Systems},
 doi = {10.52202/079017-1154},
 editor = {A. Globerson and L. Mackey and D. Belgrave and A. Fan and U. Paquet and J. Tomczak and C. Zhang},
 pages = {36602--36633},
 publisher = {Curran Associates, Inc.},
 title = {Unpacking DPO and PPO: Disentangling Best Practices for Learning from Preference Feedback},
 url = {https://proceedings.neurips.cc/paper_files/paper/2024/file/404df2480b6eef0486a1679e371894b0-Paper-Conference.pdf},
 volume = {37},
 year = {2024}
}

@inproceedings{kwon2023efficient,
author = {Kwon, Woosuk and Li, Zhuohan and Zhuang, Siyuan and Sheng, Ying and Zheng, Lianmin and Yu, Cody Hao and Gonzalez, Joseph and Zhang, Hao and Stoica, Ion},
title = {Efficient Memory Management for Large Language Model Serving with PagedAttention},
year = {2023},
isbn = {9798400702297},
publisher = {Association for Computing Machinery},
address = {New York, NY, USA},
url = {https://doi.org/10.1145/3600006.3613165},
doi = {10.1145/3600006.3613165},
booktitle = {Proceedings of the 29th Symposium on Operating Systems Principles},
pages = {611–626},
numpages = {16},
location = {Koblenz, Germany},
series = {SOSP '23}
}

@article{abdin2024phi,
  title={Phi-4 technical report},
  author={Abdin, Marah and Aneja, Jyoti and Behl, Harkirat and Bubeck, S{\'e}bastien and Eldan, Ronen and Gunasekar, Suriya and Harrison, Michael and Hewett, Russell J and Javaheripi, Mojan and Kauffmann, Piero and others},
  journal={arXiv preprint arXiv:2412.08905},
  year={2024}
}

@misc{mistral_large_2_blog,
  author = {{Mistral AI Team}},
  title  = {Large Enough: Announcement of Mistral Large 2},
  year   = {2024},
  url    = {https://mistral.ai/news/mistral-large-2407},
}

@misc{mistral_small_3_blog,
  author = {{Mistral AI Team}},
  title  = {Mistral Small 3},
  year   = {2025},
  url    = {https://mistral.ai/news/mistral-small-3},
}

@inproceedings{
singhal2025llamanemotron,
title={Llama-Nemotron: Efficient Reasoning Models},
author={Soumye Singhal and Jiaqi Zeng and Alexander Bukharin and Yian Zhang and Gerald Shen and Ameya Sunil Mahabaleshwarkar and Bilal Kartal and Yoshi Suhara and Akhiad Bercovich and Itay Levy and Izik Golan and Mohammed Dabbah and Ran El-Yaniv and Somshubra Majumdar and Igor Gitman and Evelina Bakhturina and Jimmy J. Zhang and Bor-Yiing Su and Guyue Huang and Izzy Putterman and Mostofa Patwary and Oluwatobi Olabiyi and Olivier Delalleau and Bryan Catanzaro and Boris Ginsburg and Oleksii Kuchaiev and Tugrul Konuk},
booktitle={The Exploration in AI Today Workshop at ICML 2025},
year={2025},
url={https://openreview.net/forum?id=ev1xpo9mbI}
}

@misc{liu2025muonscalablellmtraining,
      title={Muon is Scalable for LLM Training}, 
      author={Jingyuan Liu and Jianlin Su and Xingcheng Yao and Zhejun Jiang and Guokun Lai and Yulun Du and Yidao Qin and Weixin Xu and Enzhe Lu and Junjie Yan and Yanru Chen and Huabin Zheng and Yibo Liu and Shaowei Liu and Bohong Yin and Weiran He and Han Zhu and Yuzhi Wang and Jianzhou Wang and Mengnan Dong and Zheng Zhang and Yongsheng Kang and Hao Zhang and Xinran Xu and Yutao Zhang and Yuxin Wu and Xinyu Zhou and Zhilin Yang},
      year={2025},
      eprint={2502.16982},
      archivePrefix={arXiv},
      primaryClass={cs.LG},
      url={https://arxiv.org/abs/2502.16982}, 
}

@misc{cohere2025commandaenterprisereadylarge,
      title={Command A: An Enterprise-Ready Large Language Model}, 
      author={Team Cohere and Aakanksha and Arash Ahmadian and Marwan Ahmed and Jay Alammar and Yazeed Alnumay and Sophia Althammer and Arkady Arkhangorodsky and Viraat Aryabumi and Dennis Aumiller and Raphaël Avalos and Zahara Aviv and Sammie Bae and Saurabh Baji and Alexandre Barbet and Max Bartolo and Björn Bebensee and Neeral Beladia and Walter Beller-Morales and Alexandre Bérard and Andrew Berneshawi and Anna Bialas and Phil Blunsom and Matt Bobkin and Adi Bongale and Sam Braun and Maxime Brunet and Samuel Cahyawijaya and David Cairuz and Jon Ander Campos and Cassie Cao and Kris Cao and Roman Castagné and Julián Cendrero and Leila Chan Currie and Yash Chandak and Diane Chang and Giannis Chatziveroglou and Hongyu Chen and Claire Cheng and Alexis Chevalier and Justin T. Chiu and Eugene Cho and Eugene Choi and Eujeong Choi and Tim Chung and Volkan Cirik and Ana Cismaru and Pierre Clavier and Henry Conklin and Lucas Crawhall-Stein and Devon Crouse and Andres Felipe Cruz-Salinas and Ben Cyrus and Daniel D'souza and Hugo Dalla-Torre and John Dang and William Darling and Omar Darwiche Domingues and Saurabh Dash and Antoine Debugne and Théo Dehaze and Shaan Desai and Joan Devassy and Rishit Dholakia and Kyle Duffy and Ali Edalati and Ace Eldeib and Abdullah Elkady and Sarah Elsharkawy and Irem Ergün and Beyza Ermis and Marzieh Fadaee and Boyu Fan and Lucas Fayoux and Yannis Flet-Berliac and Nick Frosst and Matthias Gallé and Wojciech Galuba and Utsav Garg and Matthieu Geist and Mohammad Gheshlaghi Azar and Seraphina Goldfarb-Tarrant and Tomas Goldsack and Aidan Gomez and Victor Machado Gonzaga and Nithya Govindarajan and Manoj Govindassamy and Nathan Grinsztajn and Nikolas Gritsch and Patrick Gu and Shangmin Guo and Kilian Haefeli and Rod Hajjar and Tim Hawes and Jingyi He and Sebastian Hofstätter and Sungjin Hong and Sara Hooker and Tom Hosking and Stephanie Howe and Eric Hu and Renjie Huang and Hemant Jain and Ritika Jain and Nick Jakobi and Madeline Jenkins and JJ Jordan and Dhruti Joshi and Jason Jung and Trushant Kalyanpur and Siddhartha Rao Kamalakara and Julia Kedrzycki and Gokce Keskin and Edward Kim and Joon Kim and Wei-Yin Ko and Tom Kocmi and Michael Kozakov and Wojciech Kryściński and Arnav Kumar Jain and Komal Kumar Teru and Sander Land and Michael Lasby and Olivia Lasche and Justin Lee and Patrick Lewis and Jeffrey Li and Jonathan Li and Hangyu Lin and Acyr Locatelli and Kevin Luong and Raymond Ma and Lukas Mach and Marina Machado and Joanne Magbitang and Brenda Malacara Lopez and Aryan Mann and Kelly Marchisio and Olivia Markham and Alexandre Matton and Alex McKinney and Dominic McLoughlin and Jozef Mokry and Adrien Morisot and Autumn Moulder and Harry Moynehan and Maximilian Mozes and Vivek Muppalla and Lidiya Murakhovska and Hemangani Nagarajan and Alekhya Nandula and Hisham Nasir and Shauna Nehra and Josh Netto-Rosen and Daniel Ohashi and James Owers-Bardsley and Jason Ozuzu and Dennis Padilla and Gloria Park and Sam Passaglia and Jeremy Pekmez and Laura Penstone and Aleksandra Piktus and Case Ploeg and Andrew Poulton and Youran Qi and Shubha Raghvendra and Miguel Ramos and Ekagra Ranjan and Pierre Richemond and Cécile Robert-Michon and Aurélien Rodriguez and Sudip Roy and Laura Ruis and Louise Rust and Anubhav Sachan and Alejandro Salamanca and Kailash Karthik Saravanakumar and Isha Satyakam and Alice Schoenauer Sebag and Priyanka Sen and Sholeh Sepehri and Preethi Seshadri and Ye Shen and Tom Sherborne and Sylvie Chang Shi and Sanal Shivaprasad and Vladyslav Shmyhlo and Anirudh Shrinivason and Inna Shteinbuk and Amir Shukayev and Mathieu Simard and Ella Snyder and Ava Spataru and Victoria Spooner and Trisha Starostina and Florian Strub and Yixuan Su and Jimin Sun and Dwarak Talupuru and Eugene Tarassov and Elena Tommasone and Jennifer Tracey and Billy Trend and Evren Tumer and Ahmet Üstün and Bharat Venkitesh and David Venuto and Pat Verga and Maxime Voisin and Alex Wang and Donglu Wang and Shijian Wang and Edmond Wen and Naomi White and Jesse Willman and Marysia Winkels and Chen Xia and Jessica Xie and Minjie Xu and Bowen Yang and Tan Yi-Chern and Ivan Zhang and Zhenyu Zhao and Zhoujie Zhao},
      year={2025},
      eprint={2504.00698},
      archivePrefix={arXiv},
      primaryClass={cs.CL},
      url={https://arxiv.org/abs/2504.00698}, 
}

@article{liu2024deepseek,
  title={Deepseek-v3 technical report},
  author={Liu, Aixin and Feng, Bei and Xue, Bing and Wang, Bingxuan and Wu, Bochao and Lu, Chengda and Zhao, Chenggang and Deng, Chengqi and Zhang, Chenyu and Ruan, Chong and others},
  journal={arXiv preprint arXiv:2412.19437},
  year={2024}
}

@inproceedings{zhang-etal-2025-dorm,
    title = "{DORM}: Preference Data Weights Optimization for Reward Modeling in {LLM} Alignment",
    author = "Zhang, Rongzhi  and
      Zhang, Chenwei  and
      Zhang, Xinyang  and
      Qiu, Liang  and
      Jiang, Haoming  and
      Zhuang, Yuchen  and
      Zhang, Qingru  and
      Yun, Hyokun  and
      Li, Xian  and
      Yin, Bing  and
      Zhao, Tuo  and
      Zhang, Chao",
    editor = "Christodoulopoulos, Christos  and
      Chakraborty, Tanmoy  and
      Rose, Carolyn  and
      Peng, Violet",
    booktitle = "Findings of the Association for Computational Linguistics: EMNLP 2025",
    month = nov,
    year = "2025",
    address = "Suzhou, China",
    publisher = "Association for Computational Linguistics",
    url = "https://aclanthology.org/2025.findings-emnlp.1237/",
    doi = "10.18653/v1/2025.findings-emnlp.1237",
    pages = "22721--22739",
    ISBN = "979-8-89176-335-7"
}

@inproceedings{
liang2022rewarduncertaintyexplorationpreferencebased,
title={Reward Uncertainty for Exploration in Preference-based Reinforcement Learning},
author={Xinran Liang and Katherine Shu and Kimin Lee and Pieter Abbeel},
booktitle={International Conference on Learning Representations},
year={2022},
url={https://openreview.net/forum?id=OWZVD-l-ZrC}
}

@inproceedings{
    mehta2025sampleefficientpreferencealignment,
    title={Sample Efficient Preference Alignment in {LLM}s via Active Exploration},
    author={Viraj Mehta and Syrine Belakaria and Vikramjeet Das and Ojash Neopane and Yijia Dai and Ilija Bogunovic and Barbara E Engelhardt and Stefano Ermon and Jeff Schneider and Willie Neiswanger},
    booktitle={Second Conference on Language Modeling},
    year={2025},
    url={https://openreview.net/forum?id=Vi5cIfIslX}
}

@article{cercola2025efficientreinforcementlearninghuman,
title = {Efficient Reinforcement Learning from Human Feedback via Bayesian preference inference},
journal = {IFAC Journal of Systems and Control},
volume = {35},
pages = {100398},
year = {2026},
issn = {2468-6018},
doi = {https://doi.org/10.1016/j.ifacsc.2026.100398},
url = {https://www.sciencedirect.com/science/article/pii/S2468601826000386},
author = {Matteo Cercola and Valeria Capretti and Simone Formentin}
}

@inproceedings{
loshchilov2017decoupled,
title={Decoupled Weight Decay Regularization},
author={Ilya Loshchilov and Frank Hutter},
booktitle={International Conference on Learning Representations},
year={2019},
url={https://openreview.net/forum?id=Bkg6RiCqY7},
}

@misc{yang2026rewarduqunifiedframeworkuncertaintyaware,
      title={RewardUQ: A Unified Framework for Uncertainty-Aware Reward Models}, 
      author={Daniel Yang and Samuel Stante and Florian Redhardt and Lena Libon and Parnian Kassraie and Ido Hakimi and Barna Pásztor and Andreas Krause},
      year={2026},
      eprint={2602.24040},
      archivePrefix={arXiv},
      primaryClass={cs.LG},
      url={https://arxiv.org/abs/2602.24040}, 
}
\bibliographystyle{icml2026}

\newpage
\appendix
\onecolumn

\appendixtableofcontents
\clearpage

\makeatletter

\let\apx@orig@section\section
\RenewDocumentCommand{\section}{s o m}{%
  \IfBooleanTF{#1}{%
    \apx@orig@section*{#3}%
  }{%
    \IfValueTF{#2}{\apx@orig@section[#2]{#3}}{\apx@orig@section{#3}}%
    \edef\apx@dest{\@currentHref}%
    \apx@writetocline{section}%
      {\protect\numberline{\thesection}\IfValueTF{#2}{#2}{#3}}%
      {\thepage}%
      {\apx@dest}%
  }%
}

\let\apx@orig@subsection\subsection
\RenewDocumentCommand{\subsection}{s o m}{%
  \IfBooleanTF{#1}{%
    \apx@orig@subsection*{#3}%
  }{%
    \IfValueTF{#2}{\apx@orig@subsection[#2]{#3}}{\apx@orig@subsection{#3}}%
    \edef\apx@dest{\@currentHref}%
    \apx@writetocline{subsection}%
      {\protect\numberline{\thesubsection}\IfValueTF{#2}{#2}{#3}}%
      {\thepage}%
      {\apx@dest}%
  }%
}

\makeatother
\clearpage

\section{Author Contributions}

All authors provided valuable contributions to implementing, analyzing, and iterating on experiments, writing and the paper, and managing the overall progress of the project.

\textbf{DM} proposed and implemented the DRTS and DeltaUCB algorithms, and conducted extensive experiments evaluating response pair selection methods, generalization to diverse input prompt datasets, and comprehensive ablations on both sample efficiency and preference optimization algorithms. Furthermore, he developed and maintained major parts of the coding infrastructure, including the acquisition functions (such as MaxMinLCB and DTS), the main active learning loop for preference data generation, and the training scripts for DPO, IPO, SimPO and Reward Modeling, along with the evaluation scripts for measuring downstream and reward model performance. To scale the system, he added multi-node support for both training and the active learning loop, and optimized the overall pipeline. Additionally, \textbf{DM} experimented on and finalized the LLM-as-a-Judge setup, and conducted rigorous hyperparameter searches to ensure pipeline robustness and maximize performance across all active learning and training components. He led the release of the preference data and trained models and actively contributed to the writing of all sections of the paper.

\textbf{MS} developed and maintained major parts of the coding infrastructure, in particular the core inference pipeline used for both response generation and annotation. He curated the model pool, balancing diversity and quality, and built the surrounding infrastructure. Based on this, he conducted experiments and ablations on the model pool and judge. Furthermore, \textbf{MS} implemented evaluation pipelines for reward models, downstream benchmarks, and judge models, and contributed to the implementation of acquisition functions from prior work. He contributed to the training and efficient sweeping setup. Building on this, he conducted experiments on the effect of different judges and acquisition functions on reward models and downstream task performance, informing design choices in the final pipeline. Finally, he led the open-source release of the codebase and actively contributed to writing all sections of the paper.

\textbf{JL} worked on the response generation stage of the pipeline, contributed to the implementation of acquisition functions from existing works, and proposed the MaxMin baseline. She established schemas to ensure consistency and compliance across the data and methods, and led the open-source release of the dataset. She analysed several experimental design choices, including response quality annotation through probabilistic scoring with LLM-as-a-Judge and the validity of using different judge models for AlpacaEval 2. In addition, \textbf{JL} actively contributed to optimization and refactoring efforts, including improvements to the active learning loop, as well as to the writing of all sections of the paper and rebuttal.

\textbf{MW} contributed to the implementation and analysis of the active preference data
generation pipeline and ran reward-model and downstream evaluations. He was primarily responsible for writing the Abstract,
Introduction, Related Work, Conclusion, and the Appendix sections, and contributed to editing the remaining sections of the paper. Moreover, \textbf{MW} contributed to the rebuttal and the corresponding refactoring of the paper. 

\textbf{IH}, \textbf{BP}, and \textbf{AK} supervised the research, suggested ideas and experiments,
guided the project direction, and assisted with writing and editing the paper.

\section{Response Generation} \label{app:reponse_generation}

This section details the response generation step (\cref{sec:response_generation}) of \textsc{ActiveUltraFeedback}, in which we use vLLM~\citep{kwon2023efficient} with a large model pool of diverse LLMs to generate candidate responses for the input prompts.

\subsection{Model Pool} \label{app:model_pool}

\cref{tab:response_model_pool} lists the 30 LLMs forming our model pool. We include a wide range of both model families (12 different model families, e.g.  Qwen 2.5~\citep{qwen2025qwen25technicalreport}, Qwen 3~\citep{yang2025qwen3technicalreport}, Llama 3~\citep{grattafiori2024llama3herdmodels}, Phi 4~\citep{abdin2024phi}, Mistral Large 2~\citep{mistral_large_2_blog}, Mistral Small 3~\citep{mistral_small_3_blog}, Nemotron~\citep{wang2024helpsteer, singhal2025llamanemotron}, Gemma 3~\citep{gemmateam2024gemmaopenmodelsbased}, OLMo 2~\citep{olmo20252olmo2furious}, Tulu 3~\citep{lambert2025tulu3pushingfrontiers}, SmolLM 2~\citep{allal2025smollm2smolgoesbig}, Moonlight~\citep{liu2025muonscalablellmtraining}, Command A~\citep{cohere2025commandaenterprisereadylarge}, and DeepSeek V3~\citep{liu2024deepseek}) and model sizes (0.5B to 671B) to ensure content and quality diversity, in line with prior work~\citep{cui2024ultrafeedbackboostinglanguagemodels, lambert2025tulu3pushingfrontiers}. 

\begin{table*}[ht]
    \centering
    \caption{The 30 models used for response generation with their total number of parameters (in billions) and licenses. Separators are placed between models from different families.}
    \label{tab:response_model_pool}
    \begin{tabular}{lcl}
        \toprule
        \textbf{Model} & \textbf{\# Parameters} & \textbf{License} \\
        \midrule
        Qwen/Qwen2.5-0.5B-Instruct & 0.5B & Apache 2.0 \\
        Qwen/Qwen2.5-72B-Instruct & 72B & Qwen \\
        Qwen/Qwen3-0.6B & 0.6B & Apache 2.0 \\
        Qwen/Qwen3-1.7B & 1.7B & Apache 2.0 \\
        Qwen/Qwen3-14B & 14B & Apache 2.0 \\
        Qwen/Qwen3-30B-A3B & 30B & Apache 2.0 \\
        Qwen/Qwen3-32B & 32B & Apache 2.0 \\
        Qwen/Qwen3-235B-A22B & 235B & Apache 2.0 \\
        \midrule
        meta-llama/Llama-3.1-8B-Instruct & 8B &  Llama 3 \\
        meta-llama/Llama-3.2-1B-Instruct & 1B &  Llama 3 \\
        meta-llama/Llama-3.2-3B-Instruct & 3B &  Llama 3 \\
        meta-llama/Llama-3.3-70B-Instruct & 70B &  Llama 3 \\
        \midrule
        microsoft/Phi-4-mini-instruct & 4B & MIT \\
        microsoft/phi-4 & 14B &  MIT \\
        \midrule
        mistralai/Mistral-Small-24B-Instruct-2501 & 23B &  Apache 2.0 \\
        \midrule
        mistralai/Mistral-Large-Instruct-2411 & 123B &  MRL \\
        \midrule
        nvidia/Llama-3.1-Nemotron-70B-Instruct-HF & 70B &  Llama 3 \\
        nvidia/Llama-3\_3-Nemotron-Super-49B-v1 & 49B & Nvidia Open Model \\
        nvidia/Llama-3\_1-Nemotron-Ultra-253B-v1 & 253B & Nvidia Open Model \\
        \midrule
        google/gemma-3-1b-it & 1B &  Gemma \\
        google/gemma-3-4b-it & 4B &  Gemma \\
        google/gemma-3-12b-it & 12B &  Gemma \\
        google/gemma-3-27b-it & 27B &  Gemma \\
        \midrule
        allenai/OLMo-2-0325-32B-Instruct & 32B &  Apache 2.0 \\
        allenai/Llama-3.1-Tulu-3-70B & 70B &  Llama 3 \\
        allenai/Llama-3.1-Tulu-3-405B & 405B &  Llama 3 \\
        \midrule
        HuggingFaceT/SmolLM2-1.7B-Instruct & 1.7B & Apache 2.0\\
        \midrule
        moonshotai/Moonlight-16B-A3B-Instruct & 16B & MIT \\
        \midrule
        CohereLabs/c4ai-command-a-03-2025 & 111B &  CC by NC 4.0 \\
        \midrule
        deepseek-ai/DeepSeek-V3 & 671B &  Deepseek \\
        \bottomrule
    \end{tabular}
\end{table*}

\subsection{Response Principles} \label{app:response_principles}

Beyond model diversity (\cref{app:model_pool}), we introduce diversity through guiding principles that the LLMs should follow when generating responses. For every prompt-model pair, we uniformly sample a guiding principle among truthfulness, honesty, and helpfulness, at random following the UltraFeedback pipeline's approach~\citep{cui2024ultrafeedbackboostinglanguagemodels}. To demonstrate the principle to the model, we then randomly sample a system prompt, among 11 system prompts, for the sampled principle.
We adopt the prompt templates from the UltraFeedback pipeline but explicitly exclude the verbalized calibration principle. This modification prevents the subsequent annotation step (\cref{app:annotation}) from being biased by the model's self-expressed uncertainty, which could otherwise lead to artificially lower scores for responses where the model expresses doubt.
See \cref{app:response_generation_prompt_templates} for the system prompts.

\section{ENN Reward Model}\label{app:reward_model}

Following prior work~\citep{dwaracherla2024efficientexplorationllms, melo2024deep, liu2024sampleefficientalignmentllms}, we utilize the Epistemic Neural Network (ENN) \citep{osband2023epistemicneuralnetworks} architecture, implemented by \citep{yang2026rewarduqunifiedframeworkuncertaintyaware}, to model the reward function. Unlike standard reward models (\cref{sec:background}) that provide a single scalar point estimate, an ENN represents a distribution over reward functions, $p(r | \mathcal{D})$, where $\mathcal{D}$ is the set of observed preferences. This allows the model to quantify the epistemic uncertainty, the uncertainty stemming from a lack of data, which is the foundation for our active learning methods.

\subsection{Architecture} \label{app:reward_model_architecture}

We implement the ENN using an ensemble built on top of a fixed, pre-trained language model. This architecture consists of two components: a shared backbone and an ensemble of reward heads.

First, for any prompt-response pair $(x, y)$, we extract a feature vector $h(x, y)$ using a pre-trained LLM backbone. We utilize the embedding of the final token from the last hidden layer as the representation. Crucially, this backbone is kept frozen and unchanged during training, so only the lightweight reward heads are updated in the active learning loop, keeping in-loop training efficient and consistent with prior ENN-based RLHF work.

Second, the reward function is estimated by an ensemble of $K$ independent Multi-Layer Perceptrons (MLPs), denoted as $\{r_{\phi_k}\}_{k=1}^K$. Each head $k$ takes the embedding $h(x, y)$ as input and outputs a scalar reward. We define the final reward estimate as the mean of the ensemble predictions, $r(x, y)$, while the epistemic uncertainty is quantified by their standard deviation, $\sigma_r(x, y)$.
The epistemic uncertainty is scaled by a hyperparameter $\beta > 0$ to obtain the lower and upper bounds of the reward estimate, $\underline{r}(x,y) = r(x,y) - \beta \sigma_r(x,y)$ and $\overline{r}(x,y) = r(x,y) + \beta \sigma_r(x,y)$ respectively.
Studying how smaller backbones or ensemble sizes affect downstream acquisition quality is an important direction for future work.

\subsection{Training} \label{app:reward_model_training}

We update the ENN reward model at the end of each \activeuf{} iteration using a replay buffer $\mathcal{B} = \{(x_i, y_i^+, y_i^-)\}$ that aggregates all preference data collected thus far. We sample (without replacement) a training dataset $\mathcal{D}_{\text{train}}$ by sampling from $\mathcal{B}$ such that its size is given by $|\mathcal{D}_{\text{train}}| = \min(|\mathcal{B}|, b \cdot \rho)$, where $b$ denotes the \activeuf{} batch size and $\rho$ is a hyperparameter controlling the magnitude of $\mathcal{D}_{\text{train}}$. 

The parameters $\phi = \{\phi_k\}_{k=1}^K$ for the $K$ ensemble heads are updated on $\mathcal{D}_{\text{train}}$ by minimizing the regularized Bradley-Terry negative log-likelihood:
\begin{equation} \label{eq:enn_objective}
    \begin{aligned}
        \mathcal{J}(\phi) = \frac{1}{K} \sum_{k=1}^K & \Bigg( \mathbb{E}_{(x, y^+, y^-) \sim \mathcal{D}_{\text{train}}} \left[ -\log \operatorname{s}\left(r_{\phi_k}(x, y^+) - r_{\phi_k}(x, y^-)\right) \right] \\
        & + \gamma \mathbb{E}_{(x, y^+, y^-) \sim \mathcal{D}_{\text{train}}} \left[ (r_{\phi_k}(x, y^+) + r_{\phi_k}(x, y^-))^2 \right] \\
        & + \zeta \lVert\phi_k - \widetilde{\phi}_k\rVert_2^2 \Bigg),
    \end{aligned}
\end{equation}
where $\operatorname{s}(x) = (1+e^{-x})^{-1}$ is the sigmoid function. 
In addition to the standard Bradley-Terry objective, this objective contains two regularization terms.
The first term, controlled by $\gamma$, centers the predicted rewards around zero. Since the Bradley-Terry probability is invariant to additive constants ($s(a-b) = s((a+c)-(b+c))$), different heads can arbitrarily drift in absolute value. This term prevents such drift, ensuring that the ensemble variance reflects genuine uncertainty rather than arbitrary offsets between heads.
The second term, controlled by $\zeta$, anchors each head $k$ to its fixed, random initialization $\widetilde{\phi}_k$. This prevents the ensemble from collapsing to a single solution, thereby preserving the diversity required for uncertainty estimation. As this is most relevant during early stages of training, where gradients tend to be large, but less relevant in later stages, the $\zeta$ parameter decays exponentially over the iterations of \activeuf.
For a complete list of training hyperparameters, see \cref{app:hyperparameters}.

\section{Response Pair Selection Methods}\label{app:details_on_acquisition_functions}

This section explains the response pair selection algorithms from \cref{sec:response_pair_acquisition} in detail.
For simplicity in notation, we drop the indexing by $i$ and consider a single prompt $x$ only. Let $\{y_{j}\}_{j=1}^m$ be the responses to $x$, and denote the corresponding
lower and upper bounds of the reward estimate as vectors by  $\underline{r}$ and $\overline{r}$.

\paragraph{\textsc{InfoMax}} \citep{infobasedactiveselection} focuses purely on exploration with a goal to reduce uncertainty uniformly; therefore, it selects the ordered pair $(j,j')$ with $j\neq j'$ that maximizes the width of the confidence interval on the preference probability,
\(
\arg\max_{j\neq j'}\ \overline{p}(y_j \succ y_{j'})-\underline{p}(y_j \succ y_{j'}) ,
\)
ignoring predicted reward quality.

\begin{algorithm}[H]
\caption{\textsc{InfoMax}}
\label{alg:infomax}
\begin{algorithmic}[1]
\FUNCTION{InfoMax$(\underline{p}, \overline{p})$}
    \STATE Compute $\Delta_{j,j'}$ for all $j,j'\in\{1,\dots,m\}$
    \STATE $\displaystyle
    \Delta_{j,j'} \gets
    \begin{cases}
    -\infty, & j=j',\\
    \overline{p}(y_j \succ y_{j'})-\underline{p}(y_j \succ y_{j'}), & j\neq j'.
    \end{cases}$
    \STATE \textbf{return} $\arg\max_{(j,j')} \Delta_{j,j'}$
\ENDFUNCTION
\end{algorithmic}
\end{algorithm}

\paragraph{\textsc{Double Thompson Sampling (DTS)}} \citep{wu2016doublethompsonsamplingdueling}
balances exploration-exploitation by sampling a perturbed utility score for each response uniformly between its lower bound $\underline r$ and upper bound $\overline r$ and choosing the top response $y_j$; the second response $y_{j'}$ is obtained by resampling until $j' \neq j$ (up to \texttt{maxiter}) with a uniform-random fallback.

\begin{algorithm}[H]
\caption{\textsc{Double Thompson Sampling (DTS)}}
\begin{algorithmic}[1]
\FUNCTION{DTS$(\underline{r}, \overline{r},\text{maxiter})$}
\STATE $j \gets \textsc{ThompsonSample}(\underline{r}, \overline{r})$ \COMMENT{first draw}
\FOR{$t=1$ to $\texttt{maxiter}$}
  \STATE $j' \gets \textsc{ThompsonSample}(\underline{r}, \overline{r})$; \COMMENT{resample until distinct}
  \IF{$j \neq j'$}
    \STATE \textbf{return} $(j,j')$
  \ENDIF
\ENDFOR
\STATE \textbf{return} $(j, \mathrm{Unif}(\{1,\dots,m\} \setminus \{j\}))$ \COMMENT{fallback after $\texttt{maxiter}$ resamples}
\ENDFUNCTION
\end{algorithmic}
\end{algorithm}

\paragraph{\textsc{MaxMinLCB}} \citep{pasztor2024bandits} is based on pairwise lower confidence bounds (\cref{eq:lcb}). It selects
\(
j_1=\arg\max_j\min_{j'\neq j}\underline{p}(y_j\succ y_{j'})
\)
 to maximize the worst-case LCB against any opponent, and then
\(
j_2=\arg\min_{j\neq j_1}\underline{p}(y_{j_1}\succ y_j)
\)
to identify the opponent with the smallest LCB against $j_1$.
We use $\epsilon$ for random tie-breaking among near-equal values (within $\epsilon$).

\begin{algorithm}[H]
\caption{\textsc{MaxMinLCB}}
\begin{algorithmic}[1]
\FUNCTION{MaxMinLCB$(\underline{p},\overline{p},\epsilon)$}
\STATE $L_{j,j'} \gets
\begin{cases}
-\infty, & j=j'\\
\underline{p}(y_j\succ y_{j'}), & j\neq j'
\end{cases}
\quad \forall\, j,j'\in\{1,\dots,m\}$ \COMMENT{ignore self/filtered pairs}

\STATE $m_j\gets \min_{j\neq j'} L_{j,j'}\ \forall j$ \COMMENT{worst-case LCB for each $j$}
\STATE $j_1\gets \textsc{RandomTieBreak}\{j:\ |m_j-\max_{j'} m_{j'}|<\epsilon\}$ \COMMENT{$\epsilon$-ties on maximin}
\STATE $j_2\gets \textsc{RandomTieBreak}\{j\neq j_1:\ |L_{j_1,j}-\min_{j'\neq j_1}L_{j_1,j'}|<\epsilon\}$ \COMMENT{$\epsilon$-ties on argmin}
\STATE \textbf{return} $(j_1,j_2)$ \COMMENT{(chosen, rejected)}
\ENDFUNCTION
\end{algorithmic}
\end{algorithm}

\paragraph{\textsc{Double Reversed Thompson Sampling (DRTS)}} extends \textsc{DTS} by drawing two independent Thompson samples, uniformly between the lower bound $\underline r$ and upper bound $\overline r$ for each response, and selecting the best and worst responses under these samples, respectively.
This targets response pairs with a large expected quality gap while preserving the exploration benefits of Thompson sampling-based methods (e.g., occasionally selecting uncertain options).
The parameter \texttt{maxiter} is the maximum number of resamples used to obtain $j'\neq j$ before falling back to a uniform draw over $\{1,\dots,m\}$.

\begin{algorithm}[H]
\caption{\textsc{Double Reversed Thompson Sampling (DRTS)}}
\begin{algorithmic}[1]
\FUNCTION{DRTS$(\underline{r}, \overline{r},\text{maxiter})$}
\STATE $j\gets\textsc{ThompsonSample}(\underline{r}, \overline{r})$ \COMMENT{sampled best}
\FOR{$t=1$ to $\texttt{maxiter}$}
  \STATE ${j'}\gets\textsc{ThompsonSample}(-\overline{r}, -\underline{r})$ \COMMENT{sampled worst via reward reversal}
  \IF{$j'\neq j$}
    \STATE \textbf{return} $(j,{j'})$
  \ENDIF \COMMENT{try to ensure a distinct pair}
\ENDFOR
\STATE \textbf{return} $(j,\ \mathrm{Unif}(\{1,\dots,m\} \setminus \{j\}))$ \COMMENT{fallback after \texttt{maxiter} resamples}
\ENDFUNCTION
\end{algorithmic}
\end{algorithm}

\paragraph{\textsc{DeltaUCB}} selects an ordered response pair by maximizing the upper confidence bound on the preference probability.
Thus, \textsc{DeltaUCB} deterministically targets the most optimistically likely win under the current confidence intervals. By relying on optimistic bounds rather than stochastic sampling, \textsc{DeltaUCB} steers exploration toward pairs that could plausibly exhibit substantial quality differences under uncertainty, while remaining fully deterministic given the current confidence intervals.

\begin{algorithm}[H]
\caption{\textsc{DeltaUCB}}
\begin{algorithmic}[1]
\FUNCTION{DeltaUCB$(\overline{p})$}
\STATE $\Delta_{j,j'}\gets
\begin{cases}
-\infty,& j=j'\\
\overline{p}(y_{i,j} \succ y_{i,j'}),& j\neq j'
\end{cases}
\ \forall\, j,j'\in\{1,\dots,m\}$ \COMMENT{optimistic gap; forbid self-pairs}
\STATE \textbf{return} $\operatorname*{arg\,max}_{(j,j')} \Delta_{j,j'}$ \COMMENT{most optimistic win probability}
\ENDFUNCTION
\end{algorithmic}
\end{algorithm}

\section{Annotation}\label{app:annotation}

Given the high cost and latency of human annotation at the scale required for our experiments, we opted to use an LLM-as-a-Judge approach. Specifically, we utilize Qwen 3 235B A22B\footnote{\href{https://huggingface.co/Qwen/Qwen3-235B-A22B}{Qwen/Qwen3-235B-A22B}} to score each response. In the following, we describe how we use the LLM to score each response (\cref{app:scoring_methodology}) and ablate on the choice of Qwen 3 235B A22B, comparing it to models of different scales (\cref{app:judge_validation}).

\subsection{Scoring Methodology} \label{app:scoring_methodology}

Following recent findings~\citep{ivison2024unpacking} that per-aspect annotation is most effective for synthetic data, we utilize the aspect-wise annotation proposed in UltraFeedback~\citep{cui2024ultrafeedbackboostinglanguagemodels}, using the aspects: $\mathcal{A} = \{\text{helpfulness, truthfulness, honesty, instruction following}\}$. Specifically, we prompt our LLM-as-a-Judge for each of these aspects, using varying system prompts to guide the model to score the response for this aspect. For the full prompt templates for each aspect, we refer the reader to \cref{app:annotation_prompt_templates}.

We explicitly instruct the LLM judge to output only the raw score as a single integer between 1 and 5, strictly suppressing any reasoning or chain-of-thought text.
This strict output constraint allows us to calculate the aspect score $s_\text{aspect}$ by computing a softmax exclusively over the logits corresponding to the tokens for the digits $1$ through $5$. Given a prompt $x$, a response $y$, and the judging prompt $z_{x, y, \text{aspect}}$, the score is computed as:
\[
    s_\text{aspect}(y \mid x) = \sum_{k=1}^5 k \cdot \frac{\exp\left(\ell_k(z_{x, y, \text{aspect}})\right)}{\sum_{j=1}^5 \exp\left(\ell_j(z_{x, y, \text{aspect}})\right)}
\]
where $\ell_k(z_{x, y, \text{aspect}})$ denotes the logit output by the judge for the token corresponding to integer $k$ when given the input prompt $z_{x, y, \text{aspect}}$.

The final scalar quality score for the response is then obtained by averaging over the set of aspects:
\[
    s_{\text{overall}}(y \mid x) = \frac{1}{|\mathcal{A}|} \sum_{\text{aspect} \in \mathcal{A}} s_\text{aspect}(y \mid x).
\]

Crucially, this continuous scoring mechanism addresses the issue of score saturation. We attribute such saturation to the inherent numeric bias of LLMs, where models disproportionately favor higher integers (e.g., 5). This tendency renders competitive responses indistinguishable when using discrete labels. By utilizing the expected value over token probabilities, we capture the judge's underlying confidence, enabling fine-grained ranking even among responses with identical discrete scores.

\begin{table}[H]
    \centering
    \caption{Comparison of the four experimental judging configurations using the Qwen/Qwen3-235B-A22B model on the \textbf{UltraFeedback} dataset ($N=60'829$). Win Rate measures the percentage of samples where the judge assigned a strictly higher overall score to the preferred response. Ties occur when the calculated overall score is identical for both responses. The \textit{Probabilistic Scoring} configuration (without reasoning) was selected for the final annotation pipeline due to its superior alignment, reliability, and speed.}
    \label{tab:judge_performance}
    \begin{tabular}{l|ccc}
    \toprule
        \textbf{Configuration} & \textbf{Win Rate} & \textbf{Tie Rate} & \textbf{Parse Errors} \\
        \midrule
        \textbf{Probabilistic Scoring} & \textbf{76.70\%} & \textbf{0.0\%} & \textbf{0} \\
        Discrete Generation & 75.36\% & 14.7\% & 275 \\
        Probabilistic Scoring + Explicit Reasoning & 73.54\% & 11.3\% & 120 \\
        Discrete Generation + Explicit Reasoning & 73.37\% & 12.1\% & 20,181 \\
        \bottomrule
    \end{tabular}
\end{table}

This necessity for a distributed signal drove the decision to suppress the model's explicit reasoning capabilities. As shown in \cref{tab:judge_performance}, our experiments on the \textbf{UltraFeedback} prompts in combination with responses from our model pool (\cref{app:model_pool}) reveal that enabling reasoning degrades performance across both scoring methods. We observed that when the judge reasons, it becomes overly certain, collapsing the probability distribution over score tokens into a single peak (score saturation). In fact, the analysis confirms that with reasoning enabled, approximately 88.4\% (53'763/60'829) of the prompts resulted in a strict probability of 1.0 assigned to a single integer token for every aspect of both responses\footnote{We utilized vLLM~\citep{kwon2023efficient} for inference, configured to return the top-20 log probabilities. In these instances, only one of the target integer tokens ($1$--$5$) appeared within the top-20 candidates. This implies that the logits for the remaining score tokens were negligible, resulting in a renormalized probability of 1.0 for the top token.}. This effectively reverts the continuous signal to a discrete integer, lowering the win rate to 73.54\%. In contrast, the \textit{Probabilistic Scoring} configuration consistently maintained a distributed probability mass, avoiding collapse entirely. This preservation of uncertainty allowed this method to distinguish between competitive responses, eliminating ties and achieving a superior win rate of 76.70\%, effectively outperforming the 75.36\% achieved by the discrete generation variant.

The \textit{Probabilistic Scoring} strategy additionally encourages validity. While the \textit{Discrete Generation + Explicit Reasoning} setup suffered over 20'000 parsing failures (out of $\sim$486'000 total inference calls) due to format deviations, the selected probabilistic approach yielded zero errors across all samples. Additionally, suppressing the reasoning step resulted in a massive gain in inference throughput, operating at approximately $15\times$ the speed of the reasoning-enabled configurations ($\sim$12'000 vs. $\sim$800 samples/hr).

The parsing errors observed when scoring with Discrete Generation could have been avoided by taking the score value with highest logit. This configuration, which we call Discrete Scoring, is slightly simpler than Probabilistic Scoring, which involves taking an expectation over the score distribution. Even so, we ultimately opted for Probabilistic Scoring because we found it to perform better on RewardBench 2~\citep{malik2025rewardbench2advancingreward} -- a benchmark that has high correlation with downstream performance -- in particular on the Ties subtask for judge sensitivity to similar answers (see \cref{tab:judge_performance_rb2}). Further comparisons on the AlpacaEval 2.0 benchmark~\citep{dubois2024length} also revealed that while both scoring methodologies led to similarly high correlations with human annotations, our Probabilistic Scoring methodology achieves higher human agreement (see \cref{tab:judge_performance_ae2}). Altogether, these results support our LLM-as-a-Judge setup for the main experiments. 

\begin{table}[H]
    \centering
    \caption{Comparison of Probabilistic vs Discrete Scoring using Qwen/Qwen3-235B-A22B on RewardBench 2~\citep{malik2025rewardbench2advancingreward}.}
    \label{tab:judge_performance_rb2}
    \begin{tabular}{l|cc}
    \toprule
        \textbf{Configuration} & \textbf{Tie Rate} & \textbf{Mean} \\
        \midrule
        \textbf{Probabilistic Scoring} & \textbf{0.833} & \textbf{0.744}\\
        Discrete Scoring & 0.729 & 0.698  \\
        \bottomrule
    \end{tabular}
\end{table}

\begin{table}[H]
    \centering
    \caption{Comparison of Probabilistic vs Discrete Scoring using Qwen/Qwen3-235B-A22B on AlpacaEval 2.0~\citep{dubois2024length}.}
    \label{tab:judge_performance_ae2}
    \begin{tabular}{l|cc}
    \toprule
        \textbf{Configuration} & \textbf{Spearman's $r$} & \textbf{Human Agreement} \\
        \midrule
        \textbf{Probabilistic Scoring} & 0.67 & \textbf{62.7\%}\\
        Discrete Scoring & 0.70 & 60.4\%  \\
        \bottomrule
    \end{tabular}
\end{table}

\subsection{Judge Model Ablation} \label{app:judge_validation}

To evaluate the effectiveness of our LLM-as-a-Judge design, we evaluate our choice of judge model on RewardBench 2 \cite{malik2025rewardbench2advancingreward}. The results can be seen in \cref{tab:judge_rewardbench_eval}.

\begin{table*}[ht]
    \centering
    \caption{Rewardbench 2 scores for our judge using different models as judge models. With this comparison, we aim to cover a wide range of model sizes to examine how model size affects annotation quality. We also added Skywork-Reward-V2-Llama-3.1-8B, the current rank 1 on RewardBench 2, as a reference.}
    \begin{tabular}{l|ccccccc}
        \toprule
        \textbf{Model} & \textbf{Factuality} & \textbf{Focus} & \textbf{Math} & \textbf{Precise IF} & \textbf{Safety} & \textbf{Ties} & \textbf{Mean} \\
        \midrule
        Qwen3-32B                      & 0.787 & 0.840 & 0.710 & 0.343 & 0.844 & 0.863 & 0.731 \\
        Qwen3-235B-A22B                & 0.851 & 0.792 & 0.689 & 0.369 & 0.931 & 0.833 & 0.744 \\
        Llama-3.3-70B-Instruct         & 0.692 & 0.753 & 0.683 & 0.437 & 0.806 & 0.866 & 0.706 \\
        \midrule 
        Skywork-Reward-V2-Llama-3.1-8B & 0.844 & 0.983 & 0.770 & 0.656 & 0.967 & 0.812 & 0.839 \\
        \bottomrule
    \end{tabular}
    \label{tab:judge_rewardbench_eval}
\end{table*}

Our judge approach performs similarly for all models, yielding accurate scores. It is important to note that while Skywork-Reward-V2-Llama-3.1-8B achieves a superior score on RewardBench 2, using its rewards as annotation scores resulted in significant degradation of the fine-tuned models in our early experiments, motivating us to opt for our judge instead. Because of this, we opted to use Qwen 3 235B A22B throughout our experiments, for its strong performance for reward modeling and general fine-tuning.

\section{Implementation Details} \label{app:implementation_details}

\subsection{Evaluation Methodology} \label{app:evaluation_methodology}

To assess the quality of the datasets generated by \activeuf, we conduct experiments targeting both stages of the standard RLHF pipeline (\cref{sec:background}): reward modeling and policy optimization. By evaluating these components in isolation, we can disentangle the data's impact on both stages.
It is important to note that the models trained for evaluation are distinct from the ENN reward model utilized within the \activeuf{} acquisition loop.

For both reward modeling and fine-tuning experiments, we utilize Llama-3.1-Tulu-3-8B-SFT\footnote{\href{https://huggingface.co/allenai/Llama-3.1-Tulu-3-8B-SFT}{allenai/Llama-3.1-Tulu-3-8B-SFT}}~\citep{lambert2025tulu3pushingfrontiers} as the base model and use parameter-efficient fine-tuning via LoRA adapters~\citep{hu2022lora} and the AdamW optimizer~\citep{loshchilov2017decoupled} for all training runs.

The objectives for both trainings follow standard procedures, using the Bradley-Terry objective (\cref{eq:bradley_terry}) for reward modeling and direct preference optimization (DPO)~\citep{rafailov2023direct} for fine-tuning.

\subsection{Training Stability} \label{app:training_stability_analysis}

In this section, we analyze the stability of \activeuf{} and our evaluation setup. In order to analyse the stability of \activeuf{}, we keep the responses and annotation scores fixed, to conserve computational resources (\cref{app:compute_estimates}),  and evaluate the stability of the response pair acquisition and ENN training in \activeuf. For this, we consider two response pair selection methods. One deterministic method (\textsc{DeltaUCB}) and one sampling-based method (\textsc{DRTS}) to also evaluate the stability of sampling-based methods. The results can be seen in \cref{tab:activeuf_stability}.

\begin{table}[H]
    \centering
    \caption{Stability of \activeuf{} across 5 different random seeds with two response pair selection methods. We report the mean and standard deviation for each benchmark. Scores are reported as relative deltas to the base model.}
    \label{tab:activeuf_stability}
    \begin{tabular}{l|ccccc|c}
        \toprule \textbf{Method} & \textbf{GSM8K} & \textbf{IFEval} & \textbf{TruthfulQA}  & \textbf{AlpacaEval 2} & \textbf{Mean} & \textbf{RewardBench 2} \\
        \midrule
        \textsc{DRTS}      & $+0.057_{\pm 0.009}$ & $+0.025_{\pm 0.017}$ & $+0.132_{\pm 0.010}$ & $+0.246_{\pm 0.007}$ & $+0.114_{\pm 0.006}$ & $+0.277_{\pm 0.025}$ \\
        \textsc{DeltaUCB}  & $+0.058_{\pm 0.009}$ & $+0.017_{\pm 0.009}$ & $+0.103_{\pm 0.007}$ & $+0.230_{\pm 0.012}$ & $+0.101_{\pm 0.006}$ & $+0.282_{\pm 0.011}$ \\
        \bottomrule
    \end{tabular}
\end{table}

We observe that, for downstream evaluations, both deterministic and sampling-based methods are very stable, only having a standard deviation of $0.006$ in their mean downstream score. For reward modelling, the sampling-based methods experience slightly higher standard deviation (0.025) than the deterministic methods (0.011), which is to be expected when introducing more stochasticity through sampling.

Now we analyse the stability of our evaluation setup, starting with the DPO training. We utilize the decontaminated version of the \textbf{UltraFeedback} dataset\footnote{\href{https://huggingface.co/datasets/allenai/ultrafeedback_binarized_cleaned}{allenai/ultrafeedback\_binarized\_cleaned}}~\citep{cui2024ultrafeedbackboostinglanguagemodels} for these experiments. First, we examine the sensitivity to initialization by training with 5 different random seeds while keeping all other hyperparameters fixed. We ensure reproducibility by fixing the random seed and explicitly shuffling the dataset according to the seed before training.

As shown in \cref{tab:seed_stability}, the standard deviation across seeds is minimal ($\approx 0.003$ for the overall score), with TruthfulQA exhibiting the highest stability ($0.001$) and AlpacaEval 2 showing slightly higher variance ($0.006$), likely due to the inherent noise in generation-based evaluation.

\begin{table}[H]
    \centering
    \caption{Training stability across 5 different random seeds. We report the mean and standard deviation for each benchmark. Scores are reported as relative deltas to the base model.}
    \label{tab:seed_stability}
    \begin{tabular}{l|ccccc}
        \toprule
        \textbf{Metric} & \textbf{GSM8K} & \textbf{IFEval} & \textbf{TruthfulQA} & \textbf{AlpacaEval 2} & \textbf{Mean} \\
        \midrule
        Mean & +0.039 & +0.020 & +0.056 & +0.028 & +0.035 \\
        Std. Dev. & 0.005 & 0.006 & 0.001 & 0.006 & 0.003 \\
        \bottomrule
    \end{tabular}
\end{table}

Next, to assess the inherent randomness caused by system-level non-determinism (e.g., PyTorch non-determinism, and non-associativity of rounding operations for floating-point numbers in multi-GPU setups), we performed 5 independent training runs using a fixed seed of 42. The results in \cref{tab:system_stability} confirm that system-level noise produces deviations comparable to seed variation ($\approx 0.004$ overall). IFEval shows slightly higher variance here ($0.011$), while TruthfulQA remains perfectly stable.

\begin{table}[H]
    \centering
    \caption{Training stability across 5 runs with a fixed seed (Seed 42), assessing system-level non-determinism. Scores are reported as relative deltas to the base model.}
    \label{tab:system_stability}
    \begin{tabular}{l|ccccc}
        \toprule
        \textbf{Metric} & \textbf{GSM8K} & \textbf{IFEval} & \textbf{TruthfulQA} & \textbf{AlpacaEval 2} & \textbf{Mean} \\
        \midrule
        Mean & +0.044 & +0.020 & +0.054 & +0.030 & +0.035 \\
        Std. Dev. & 0.003 & 0.011 & 0.000 & 0.008 & 0.004 \\
        \bottomrule
    \end{tabular}
\end{table}

We performed the same stability analysis for our Reward Model training using RewardBench 2. First, examining initialization sensitivity across 5 random seeds (\cref{tab:rm_seed_stability}), we observe moderate stability overall ($\approx 0.011$). However, the Ties metric exhibits significant variance ($0.072$), indicating that the model's ability to resolve subtle preference differences is highly sensitive to random initialization conditions.

\begin{table}[H]
    \centering
    \caption{Reward Model training stability across 5 different random seeds. Scores are reported as relative deltas to the base model.}
    \label{tab:rm_seed_stability}
    \begin{tabular}{l|ccccccc}
        \toprule
        \textbf{Metric} & \textbf{Factuality} & \textbf{Focus} & \textbf{Math} & \textbf{Precise IF} & \textbf{Safety} & \textbf{Ties} & \textbf{Mean} \\
        \midrule
        Mean & +0.344 & +0.495 & +0.145 & +0.095 & +0.453 & +0.253 & +0.298 \\
        Std. Dev. & 0.019 & 0.029 & 0.030 & 0.031 & 0.036 & 0.072 & 0.011 \\
        \bottomrule
    \end{tabular}
\end{table}

Second, we performed 5 independent training runs using a fixed seed of 42. The results in Table~\ref{tab:rm_system_stability} reveal negligible noise ($\approx 0.004$). Notably, the Ties variance drops to 0.008, confirming that the higher instability observed previously stems from algorithmic randomness (e.g., weight initialization, data permutation) rather than hardware-level non-determinism.

\begin{table}[H]
    \centering
    \caption{Reward Model stability across 5 runs with a fixed seed (Seed 42). Scores are reported as relative deltas to the base model.}
    \label{tab:rm_system_stability}
    \begin{tabular}{l|ccccccc}
        \toprule
        \textbf{Metric} & \textbf{Factuality} & \textbf{Focus} & \textbf{Math} & \textbf{Precise IF} & \textbf{Safety} & \textbf{Ties} & \textbf{Mean} \\
        \midrule
        Mean & +0.363 & +0.444 & +0.145 & +0.128 & +0.546 & +0.252 & +0.292 \\
        Std. Dev. & 0.005 & 0.006 & 0.007 & 0.007 & 0.006 & 0.008 & 0.004 \\
        \bottomrule
    \end{tabular}
\end{table}

Finally, we extend our stability analysis to the optimization algorithms themselves. To ensure that our performance gains are robust and not artifacts of initialization, we trained both IPO and SimPO models using 5 different random seeds. As detailed in Tables~\ref{tab:simpo_stability_checks} and~\ref{tab:ipo_stability_checks}, our setup proves to be highly stable across different preference optimization algorithms. Both methods demonstrate minimal variance across key benchmarks (e.g., standard deviations of $\approx 0.004$--$0.011$ on GSM8K and $\approx 0.005$--$0.006$ on TruthfulQA). These results, reflected in the low variance of the aggregated mean scores ($0.015$ for SimPO and $0.011$ for IPO), confirm that the improvements over the baseline are reliable and consistent.

\begin{table}[H]
    \centering
    \caption{Stability analysis of our SimPO algorithms setup. We report the Mean and Standard Deviation across 5 different random seeds. Scores are reported as relative deltas to the base model.}
    \label{tab:simpo_stability_checks}
    \begin{tabular}{l|ccccc}
    \toprule
    \textbf{Benchmark} & \textbf{GSM8K} & \textbf{IFEval} & \textbf{TruthfulQA} & \textbf{AlpacaEval} & \textbf{Mean} \\
    \midrule
    Mean Delta & +0.033 & +0.019 & +0.058 & +0.273  & +0.095 \\
    Std. Dev.  & 0.011 & 0.009 & 0.006 & 0.036  & 0.015 \\
    \bottomrule
    \end{tabular}
\end{table}

\begin{table}[H]
    \centering
    \caption{Stability analysis of our IPO algorithms setup. We report the Mean and Standard Deviation across 5 different random seeds. Scores are reported as relative deltas to the base model.}
    \label{tab:ipo_stability_checks}
    \begin{tabular}{l|ccccc}
    \toprule
    \textbf{Benchmark}  & \textbf{GSM8K} & \textbf{IFEval} & \textbf{TruthfulQA} & \textbf{AlpacaEval} & \textbf{Mean} \\
    \midrule
    Mean Delta  & +0.048 & +0.035 & +0.040 & +0.304 & +0.106 \\
    Std. Dev. & 0.004 & 0.005 & 0.005 & 0.036  & 0.011 \\
    \bottomrule
    \end{tabular}
\end{table}

\subsection{Hyperparameters}\label{app:hyperparameters}

Throughout our work, we have conducted extensive experiments for identifying well-performing and robust hyperparameters for different modules of our pipeline, including: response generation, annotation pipeline, ENN reward model, several direct preference optimization algorithms, and reward model training. In this section, we detail all hyperparameters along with their final values, and, if applicable, the sweep range we used to identify the final values.

\paragraph{Batch Size} The number of prompts per iteration of \activeuf{} is fixed at 64 for all experiments.

\paragraph{Response Generation and Annotation}
We use vLLM~\citep{kwon2023efficient} for prompting LLMs in two stages of the \activeuf{} pipeline: Response Generation (\cref{sec:response_generation}) and Preference Annotation (\cref{sec:oracle_preference_annotation}). The sampling parameters used for each stage are listed in \cref{tab:sampling_parameters}.

\begin{table}[htb]
    \centering
    \caption{Sampling parameters for Response Generation and Preference Annotation in \activeuf.}
    \begin{tabular}{l|cc}
        \toprule
        \textbf{Hyperparameter} & \textbf{Response Generation} & \textbf{Preference Annotation}  \\
        \midrule
        Temperature & 1.0 & 0.0 \\
        Top-$p$ & 1.0 & -- \\
        Max Response Tokens & 4096 & 16 \\
        \bottomrule
    \end{tabular}
    \label{tab:sampling_parameters}
\end{table}

\paragraph{ENN Reward Model}
The hyperparameters for the ENN reward model in the Reward Prediction stage of \activeuf{} (\cref{sec:reward_prediction}) are listed in \cref{tab:enn_architecture_hp}. Most values are adopted from prior work \citep{dwaracherla2024efficientexplorationllms}. As a base model for the ENN reward model, we use Skywork Reward V2 Qwen3 4B\footnote{\href{https://huggingface.co/Skywork/Skywork-Reward-V2-Qwen3-4B}{Skywork/Skywork-Reward-V2-Qwen3-4B}} for its strong reward modelling performance, and train the MLP head ensemble on the last-layer embedding of the last token in the sequence.

\begin{table}[ht]
    \centering
    \caption{Hyperparameters for the ENN architecture.}
    \begin{tabular}{l|c}
        \toprule
        \textbf{Hyperparameter} & \textbf{Value} \\
        \midrule
        Number of MLP heads & 20 \\
        Number of layers per MLP head & 2 \\
        Hidden size of each MLP head & 128 \\
        \bottomrule
    \end{tabular}
    \label{tab:enn_architecture_hp}
\end{table}

\paragraph{ENN Training}

The Reward Model Training component of \activeuf{} (\cref{sec:reward_model_training}) involves many hyperparameters. We list the ones that are fixed across all experiments in \cref{tab:hyperparameters}.

\begin{table}[H]
    \centering
    \caption{Fixed hyperparameters used across experiments for ENN training.}
    \begin{tabular}{l|c}
        \toprule
        \textbf{Hyperparameter} & \textbf{Value} \\
        \midrule
        \, Max Length (Prompt + Response) & 4096 \\
        \, Batch Size, $\lvert \mathcal{B} \rvert$ & 64 \\
        \, Train Steps & 100 \\
        \, Initial Regularization, $\zeta$ & 1.0 \\
        \, Reward Centering Coefficient, $\gamma$ & 0.01 \\
        \, Learning Rate & $5 \times 10^{-5}$ \\
        \bottomrule
    \end{tabular}
    \label{tab:hyperparameters}
\end{table}

For certain hyperparameters, the optimal value differs based on the active response pair selection method, as well as between DPO fine-tuning and reward modeling. We report the sweep performed and the optimal configuration we found in \cref{tab:acquisition_hp_sweep}.
The sweep budget was identical across all ENN-based active selection methods. The static baselines (\textsc{Random},\textsc{UltraFeedback}, \textsc{MaxMin}) do not utilize the ENN and are therefore excluded from this ENN-specific sweep.

\begin{table}[htb]
    \centering
    \caption{ENN training hyperparameters and sweep ranges for each active response pair selection method. 
    Separate optimal values were chosen based on performance after DPO fine-tuning and on RewardBench 2.}
    \begin{tabular}{l|c|ccccc}
        \toprule
        \textbf{Hyperparameter} & \textbf{Grid Values} & \textsc{InfoMax} & \textsc{DTS} & \textsc{MaxMinLCB} & \textsc{DRTS} & \textsc{DeltaUCB}  \\
        \midrule
        \multicolumn{7}{c}{\textbf{Optimal for DPO Fine-Tuning}} \\
        \midrule
        Beta $\beta$                                & [1, 2]             & 2 & 1 & 1 & 1 & 2 \\
        Regularization Decay                & [0.9, 0.99, 0.999] & 0.99 & 0.99 & 0.99 & 0.999 & 0.999\\
        Replay Buffer Size Factor, $\rho$    & [100, 1000]        & 1000 & 1000 & 1000 & 1000 & 1000 \\
        \midrule
        \multicolumn{7}{c}{\textbf{Optimal for Reward Modeling}} \\
        \midrule
        Beta $\beta$                                & [1, 2]             & 2 & 1 & 2 & 1 & 1 \\
        Regularization Decay                & [0.9, 0.99, 0.999] & 0.99 & 0.999 & 0.9 & 0.9 & 0.99 \\
        Replay Buffer Size Factor, $\rho$    & [100, 1000]        & 1000 & 1000 & 1000 & 1000 & 100 \\
        \bottomrule
    \end{tabular}
    \label{tab:acquisition_hp_sweep}
\end{table}

\paragraph{Preference Optimization (DPO, IPO, SimPO)}

To establish the optimal configuration for preference fine-tuning, we utilized the UltraFeedback dataset\footnote{\href{https://huggingface.co/datasets/allenai/ultrafeedback_binarized_cleaned}{allenai/ultrafeedback\_binarized\_cleaned}}~\citep{cui2024ultrafeedbackboostinglanguagemodels}. We conducted a hyperparameter sweep for DPO, IPO, and SimPO and selected based on best performance in our evaluation framework (\cref{app:evaluation_methodology}), are presented in \cref{tab:dpo_hyperparameters}. We fixed the batch size to 32, used a linear learning rate schedule with a warmup ratio of $0.1$, and used a max length (prompt + completion) of $2048$ for all three preference optimization algorithms.

\begin{table}[htb]
    \centering
    \caption{Optimal hyperparameters for our DPO, IPO, and SimPO fine-tuning, selected based on evaluation performance.}
    \begin{tabular}{l|c|c}
        \toprule
        \textbf{Hyperparameter} & \textbf{Grid Values} & \textbf{Chosen Value}  \\
        \midrule
        \multicolumn{3}{c}{\textbf{For DPO}} \\
        \midrule
        Learning Rate &       [$1\times10^{-6}$, $2\times10^{-5}$, $5\times10^{-4}$]                  & $2\times10^{-5}$  \\
        Lambda $\lambda$ &              [0.1, 0.01]                    & 0.1 \\
        Epochs &           [1, 3]                     & 3 \\
        \midrule 
        \multicolumn{3}{c}{\textbf{For IPO}} \\
        \midrule
        Learning Rate       & [$5\times10^{-6}$, $1\times10^{-5}$, $2\times10^{-5}$, $5\times10^{-5}$]            & $5\times10^{-6}$ \\
        Lambda $\lambda$                & [0.01, 0.1, 0.5, 1.0]            & 0.01 \\
        Epochs              & [1, 3]                           & 1 \\
        \midrule
        \multicolumn{3}{c}{\textbf{For SimPO}} \\
        \midrule
        Learning Rate       & [$5\times10^{-6}$, $1\times10^{-5}$, $2\times10^{-5}$, $5\times10^{-5}$]            & $5\times10^{-6}$ \\
        Gamma               & [0.3, 0.5, 1.0, 1.2, 1.4, 1.6]            & 1.2 \\
        Lambda $\lambda$                & [2.0, 2.5]            & 2.0 \\
        Epochs              & [1, 3]               & 1 \\
        \bottomrule
    \end{tabular}
    \label{tab:dpo_hyperparameters}
\end{table}

\paragraph{Reward Modeling}

The hyperparameter sweep and final values for reward model training, selected based on the highest mean score on RewardBench 2, are listed in \cref{tab:rm_hyperparameters}. We fixed the \textsc{Batch Size} to $128$, used a constant learning rate, and used a max length (prompt + completion) of $4096$.

\begin{table}[htb]
    \centering
    \caption{Optimal hyperparameters for reward model training, selected based on RewardBench 2 performance.}
    \begin{tabular}{l|c|c}
        \toprule
        \textbf{Hyperparameter} & \textbf{Grid Values} & \textbf{Chosen Value}  \\
        \midrule
        Learning Rate                  &  [$3\times10^{-6}$, $5\times10^{-6}$, $2\times10^{-5}$]      & $2\times10^{-5}$ \\
        Epochs                       & [1, 2, 3]         &  2 \\
        \bottomrule
    \end{tabular}
    \label{tab:rm_hyperparameters}
\end{table}

\paragraph{LoRA}
We use the hyperparameters in \cref{tab:lora_hyperparameters} for LoRA when fine-tuning (DPO, IPO, SimPO) and reward modeling.

\begin{table}[htb]
    \centering
    \caption{Hyperparameters for our LoRA setup.}
    \begin{tabular}{l|c}
        \toprule
        \textbf{Hyperparameters} & \textbf{Chosen Value}  \\
        \midrule
        Rank & 64 \\
        Alpha & 16 \\
        Dropout & 0.1 \\
        Target Modules & all-linear \\
        \bottomrule
    \end{tabular}
    \label{tab:lora_hyperparameters}
\end{table}

\subsection{Compute Estimates}\label{app:compute_estimates}

All experiments were conducted on 8 NVIDIA GH200 Grace Hopper Superchips. To facilitate extensive ablation studies and rapid iteration, we decoupled the computationally expensive generation and annotation phases from the active learning loop. Specifically, we pre-computed the candidate responses and their corresponding judge annotations for the entire dataset prior to simulating the acquisition process.

\begin{table}[ht]
    \centering
    \begin{tabular}{l|c}
        \toprule
        \textbf{Step} & \textbf{Estimated GPU Hours}  \\
        \midrule
        Response Generation    & 600 \\  

        Annotation             & 600 \\

        Active Learning Loop & 32 \\
        \bottomrule
    \end{tabular}
    \caption{Compute estimates for each step of \activeuf, estimated in GPU hours.}
    \label{tab:compute_estimate}
\end{table}

Table~\ref{tab:compute_estimate} provides a breakdown of the estimated GPU hours required for each stage of the pipeline on the \textbf{UltraFeedback} dataset. As shown, the computational budget is roughly evenly distributed between response generation and the pre-computation of judge scores. In practical use of \activeuf, the annotation cost would be drastically reduced, as the pipeline only requires annotations for the selected responses, rather than the entire candidate pool.

It is important to note that our implementation prioritized experimental flexibility and reproducibility over maximum computational efficiency. Compared to static pipelines such as \textsc{UltraFeedback}, our current implementation trades higher upfront response-generation compute for improved sample efficiency and lower annotation demand, since we pre-computed candidate responses and judge annotations for the full dataset before simulating acquisition. In practical deployment, a strong selection strategy would require processing substantially fewer selected prompts end-to-end. Moreover, \cref{app:pool_ablation} shows that competitive performance can be retained with smaller yet more diverse response model pools, suggesting a concrete path to reducing response generation cost. Further runtime reductions are also likely through more optimized distributed inference and training configurations. In total, all experiments, including model fine-tuning, reward model training, ablations, stability analyses, failed experiments, and preliminary experiments, consumed approximately 200'000 GPU hours.

\section{Additional Results}\label{app:additional_results}

\subsection{Generated Dataset Analysis} \label{app:generated_dataset_analysis}

To understand the selection dynamics of different response pair acquisition methods, we analyze the distributions of the generated datasets by examining how often each model from our pool was selected, how often it was annotated as chosen and rejected (\cref{fig:model_distribution_comparison_0}) and the mean scores for the chosen and rejected responses for different response pair selection methods (\cref{tab:mean_chosen_rejected_score_by_method}). 

We find that methods aiming at regret minimization, such as \textsc{DTS} (\cref{fig:dts_model_distribution}) and \textsc{MaxMinLCB} (\cref{fig:maxminlcb_model_distribution}), successfully identify high-quality models, with high judge scores (\cref{tab:mean_chosen_rejected_score_by_method}), resulting in distributions heavily skewed towards recent, large-scale models.
This highlights the key distinction between identifying strong responses and creating informative preference pairs: for downstream preference learning, high-quality chosen responses are not sufficient on their own if the rejected responses are also too strong, since this reduces the effective supervision signal.
In contrast, as expected, \textsc{Random} (\cref{fig:random_model_distribution}) exhibits a nearly uniform distribution, while \textsc{UltraFeedback} (\cref{fig:ultrafeedback_model_distribution}) displays a slight skew towards higher-quality models due to its "best-of-$N$" heuristic. 
Conversely, the entropy-minimizing \textsc{InfoMax} (\cref{fig:infomax_model_distribution}) disproportionately selects smaller, older models. We attribute this to the fact that recent, large-scale models consistently achieve near-perfect scores, leading to high certainty in their high quality. In contrast, smaller models exhibit erratic behaviour, occasionally producing high-scoring responses but frequently failing. This unpredictability results in higher epistemic uncertainty, driving the method to sample from them more frequently.
Finally, our proposed quality delta maximization methods, \textsc{DRTS} (\cref{fig:drts_model_distribution}) and \textsc{DeltaUCB}, produce distributions closely mirroring the high-scoring, but inefficient \textsc{MaxMin} baseline (\cref{fig:maxmin_model_distribution}), prioritizing the best and worst responses, yet achieve this efficiently by requiring only two annotations per prompt compared to \textsc{MaxMin}'s annotation of the full candidate set.

\begin{table}[H]
    \centering
    \caption{Mean score of the chosen, rejected, and overall responses from different response pair selection methods on the \textbf{UltraFeedback} prompts.}
    \label{tab:mean_chosen_rejected_score_by_method}
    \begin{tabular}{l|ccc}
        \toprule
        \textbf{Method} & \textbf{Mean Chosen Score} & \textbf{Mean Rejected Score} & \textbf{Mean Score}  \\
        \midrule
        \textsc{Random}        & 4.522 & 3.564 & 4.043 \\ 
        \textsc{UltraFeedback} & 4.747 & 3.810 & 4.279 \\ 
        \textsc{MaxMin}        & 4.925 & 1.605 & 3.625 \\ 
        \textsc{DeltaQwen}     & 4.549 & 2.924 & 3.736 \\
        \midrule
        \textsc{InfoMax}       & 3.666 & 3.156 & 3.411 \\
        \textsc{DTS}           & 4.855 & 4.584 & 4.720 \\
        \textsc{MaxMinLCB}     & 4.864 & 4.683 & 4.773 \\
        \midrule
        \textsc{DRTS}          & 4.752 & 1.968 & 3.360 \\
        \textsc{DeltaUCB}      & 4.705 & 2.113 & 3.409 \\
        \bottomrule
    \end{tabular}
\end{table}

\newcommand{\modeldistcaption}{Comparison between the number of times each model from our model pool (\cref{app:model_pool}) has been selected as chosen and rejected model on the \textbf{UltraFeedback} prompts for all response pair selection methods we consider.}
\newcommand{\modeldistsubcaption}[1]{\textbf{\textsc{#1}}: Model distribution of how often each model in our model pool has been selected by the \textsc{#1} response pair selection method. We further split this data into the number of times each model has been annotated as chosen (green) and rejected (red). Models are sorted based on the number of times they have been annotated as chosen.}
\begin{figure*}[ht]
    \centering
    
    \begin{subfigure}[b]{\textwidth}
        \centering
        \includegraphics{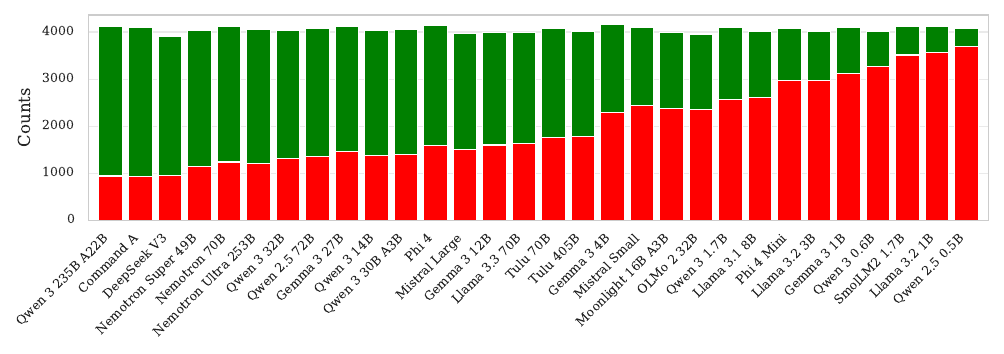}
        \caption{\modeldistsubcaption{Random}}
        \label{fig:random_model_distribution}
    \end{subfigure}

    \begin{subfigure}[b]{\textwidth}
        \centering
        \includegraphics{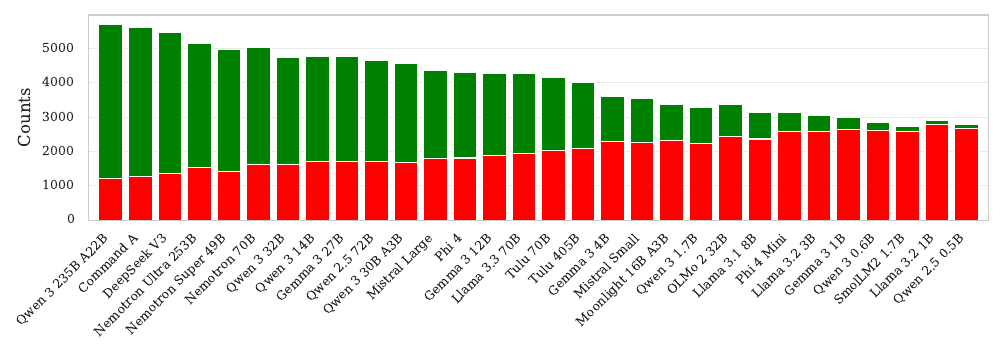}
        \caption{\modeldistsubcaption{UltraFeedback}}
        \label{fig:ultrafeedback_model_distribution}
    \end{subfigure}

    \begin{subfigure}[b]{\textwidth}
        \centering
        \includegraphics{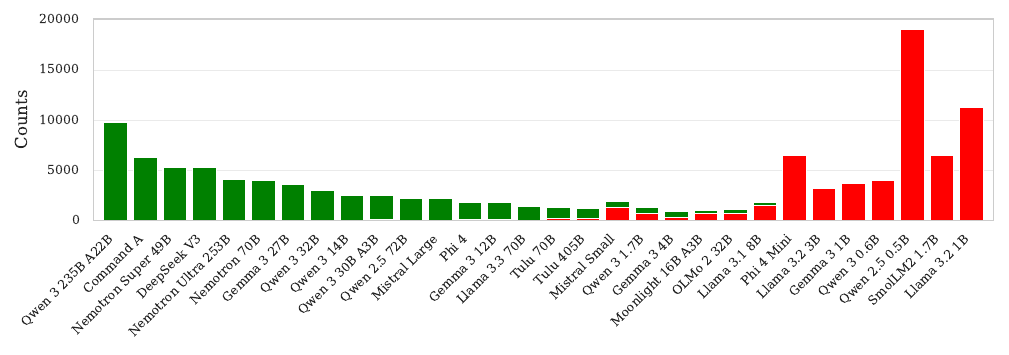}
        \caption{\modeldistsubcaption{MaxMin}}
        \label{fig:maxmin_model_distribution}
    \end{subfigure}

    \caption{\modeldistcaption}
    \label{fig:model_distribution_comparison_0}
\end{figure*}
\begin{figure*}[ht]\ContinuedFloat
    \centering
    
    \begin{subfigure}[b]{\textwidth}
        \centering
        \includegraphics{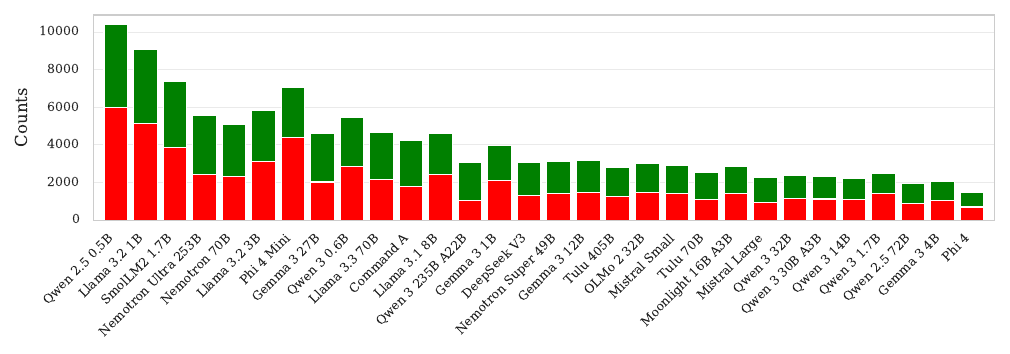}
        \caption{\modeldistsubcaption{InfoMax}}
        \label{fig:infomax_model_distribution}
    \end{subfigure}

    \begin{subfigure}[b]{\textwidth}
        \centering
        \includegraphics{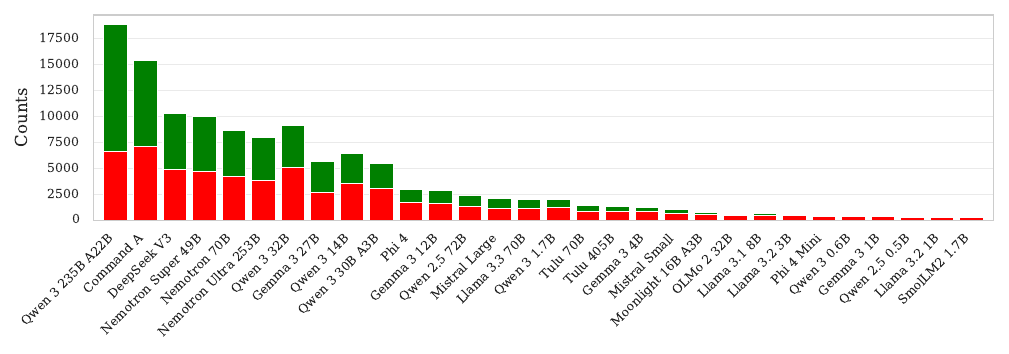}
        \caption{\modeldistsubcaption{DTS}}
        \label{fig:dts_model_distribution}
    \end{subfigure}

    \begin{subfigure}[b]{\textwidth}
        \centering
        \includegraphics{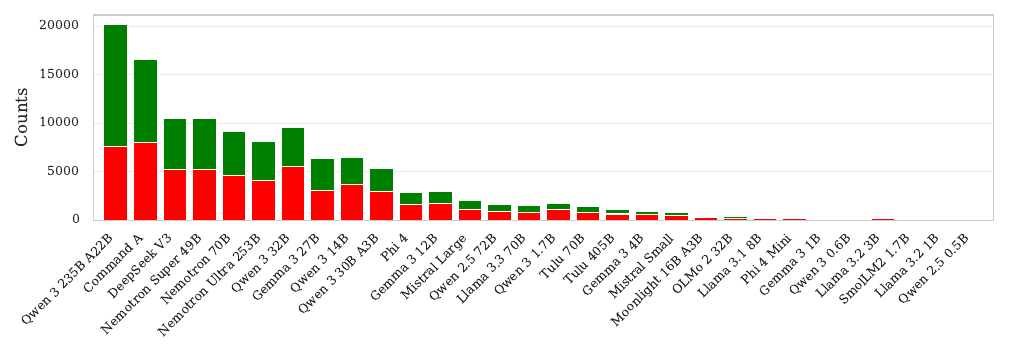}
        \caption{\modeldistsubcaption{MaxMinLCB}}
        \label{fig:maxminlcb_model_distribution}
    \end{subfigure}

    \caption{\modeldistcaption}
    \label{fig:model_distribution_comparison_1}
\end{figure*}
\begin{figure*}[ht]\ContinuedFloat
    \centering
    
    \begin{subfigure}[b]{\textwidth}
        \centering
        \includegraphics{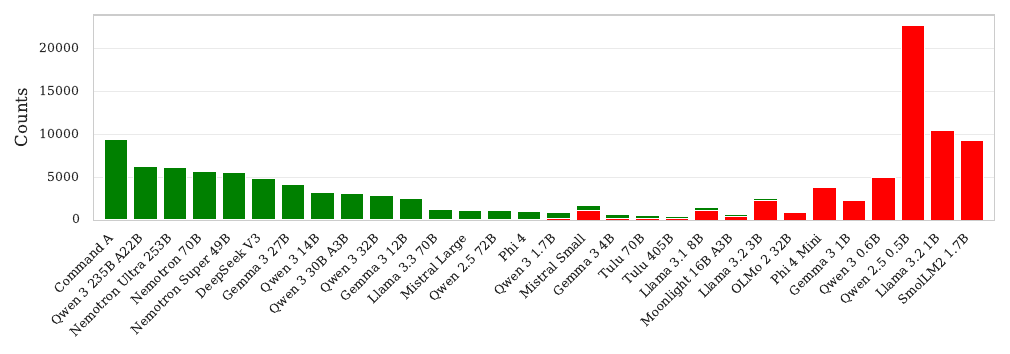}
        \caption{\modeldistsubcaption{DRTS}}
        \label{fig:drts_model_distribution}
    \end{subfigure}

    \begin{subfigure}[b]{\textwidth}
        \centering
        \includegraphics{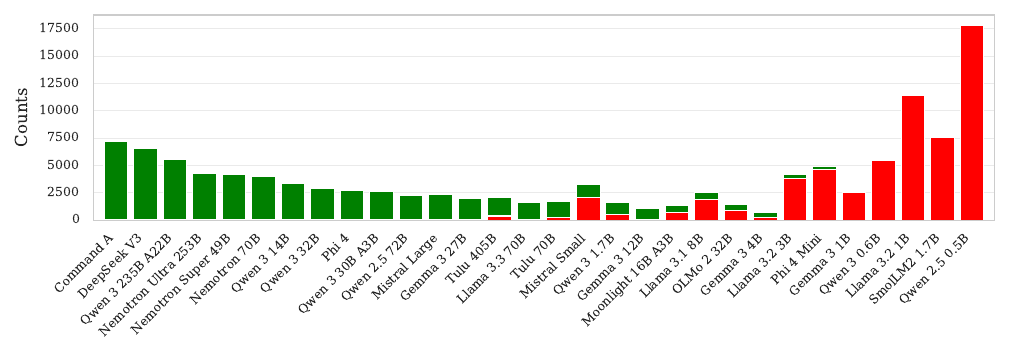}
        \caption{\modeldistsubcaption{DeltaUCB}}
        \label{fig:deltaucb_model_distribution}
    \end{subfigure}

    \caption{\modeldistcaption}
    \label{fig:model_distribution_comparison_2}
\end{figure*}

\subsection{Sample Efficiency without AlpacaEval 2} \label{app:sample_efficiency_without_alpacaeval}

\begin{figure*}[htb]
    \centering
    \includegraphics{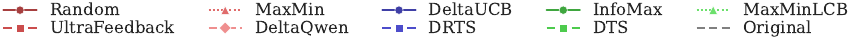}

    \begin{subfigure}[b]{0.5\textwidth}
        \centering
        \includegraphics[trim={0 0 0 0}, clip]{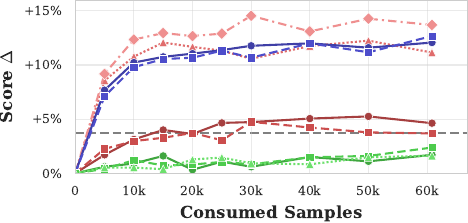}
        \caption{With AlpacaEval 2}
        \label{fig:sample_efficiency_without_alpacaeval_with}
    \end{subfigure}%
    \hfill 
    \begin{subfigure}[b]{0.5\textwidth}
        \centering
        \includegraphics[trim={0 0 0 0}, clip]{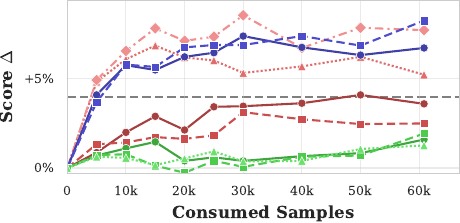}
        \caption{Without AlpacaEval 2}
        \label{fig:sample_efficiency_without_alpacaeval_without}
    \end{subfigure}
    
    \caption{Results for the sample efficiency experiment (\cref{sec:sample_efficiency_evaluation}). We compare the aggregate scores with and without AlpacaEval 2 to demonstrate how its larger magnitude dominates the mean across all benchmarks.}
    \label{fig:sample_efficiency_without_alpacaeval}
\end{figure*}

The score deltas in AlpacaEval 2 are an order of magnitude larger than those in our other benchmarks. Consequently, the mean score delta is disproportionately influenced by AlpacaEval 2, obscuring performance trends in the wider suite. To provide a clearer visualization of our sample efficiency experiment (\cref{sec:sample_efficiency_evaluation}), \cref{fig:sample_efficiency_without_alpacaeval} presents the mean performance trajectories both with and without the inclusion of AlpacaEval 2.

\subsection{Full Input Prompt Dataset Ablation}
\label{app:full_input_prompt_dataset_ablation}

In this section, we provide the detailed scores for our prompt dataset ablation (\cref{sec:input_prompt_dataset_ablation}). The detailed results, for each individual benchmark and response pair selection method, can be seen in \cref{tab:prompt_dataset_ablation}.

\begin{table*}[ht]
    \centering
    \caption{Results of \activeuf{} on downstream and reward model benchmarks using different prompt input datasets and response pair selection methods. All scores are given as relative deltas to the base model's scores for readability. Best scores are in bold. We furthermore show the scores obtained by training on the actual \textbf{UltraFeedback}, \textbf{Skywork}, and \textbf{Tulu 3} preference mixture datasets.}
    \begin{tabular}{l|ccccc|c}
        \toprule
        \textbf{Method} & \textbf{GSM8K} & \textbf{IFEval} & \textbf{TruthfulQA} & \textbf{AlpacaEval 2} & \textbf{Mean} & \textbf{RewardBench 2} \\
        \midrule
        Base Model & \phantom{+}0.758 & \phantom{+}0.713 & \phantom{+}0.468 & \phantom{+}0.083 & \phantom{+}0.506 & \phantom{+}0.290 \\
        \midrule
        \multicolumn{7}{c}{\textbf{UltraFeedback Prompts}} \\
        \midrule
        Original & +0.039 & +0.025 & +0.055 & +0.030 & +0.037 & +0.295 \\
        \textsc{Random} & +0.024 & +0.028 & +0.056 & +0.077 & +0.046 & +0.278 \\
        \textsc{UltraFeedback} & +0.037 & -0.001 & +0.039 & +0.072 & +0.036 & +0.287 \\
        \textsc{MaxMin} & +0.022 & -0.016 & \textbf{+0.150} & +0.289 & +0.111 & +0.318 \\
        \textsc{DeltaQwen} & \textbf{+0.055} & +0.047 & +0.130 & \textbf{+0.316} & \textbf{+0.137} & +0.100 \\
        \textsc{InfoMax} & +0.011 & +0.019 & +0.018 & +0.020 & +0.016 & +0.297 \\
        \textsc{DTS} & +0.011 & +0.034 & +0.013 & +0.037 & +0.023 & +0.224 \\
        \textsc{MaxMinLCB} & +0.015 & +0.017 & +0.006 & +0.027 & +0.016 & +0.230 \\
        \textsc{DRTS} & \textbf{+0.055} & \textbf{+0.050} & +0.143 & +0.259 & +0.127 & +0.312 \\
        \textsc{DeltaUCB} & +0.040 & +0.025 & +0.137 & +0.281 & +0.120 & \textbf{+0.339} \\
        \midrule
        \multicolumn{7}{c}{\textbf{Skywork Prompts}} \\
        \midrule
        Original & +0.008 & +0.052 & +0.048 & +0.066 & +0.044 & \textbf{+0.377} \\
        \textsc{Random} & +0.012 & +0.015 & +0.045 & +0.063 & +0.033 & +0.223 \\
        \textsc{UltraFeedback} & +0.027 & \textbf{+0.054} & +0.043 & +0.071 & +0.048 & +0.234 \\
        \textsc{MaxMin} & +0.049 & -0.011 & +0.128 & +0.270 & +0.108 & +0.325 \\
        \textsc{DeltaQwen} & \textbf{+0.058} & +0.002 & \textbf{+0.152} & \textbf{+0.384} & \textbf{+0.149} & +0.129 \\
        \textsc{InfoMax} & +0.021 & +0.002 & +0.011 & +0.013 & +0.012 & +0.244 \\
        \textsc{DTS} & +0.008 & +0.002 & +0.011 & +0.021 & +0.010 & +0.219 \\
        \textsc{MaxMinLCB} & +0.003 & +0.010 & +0.004 & +0.018 & +0.008 & +0.184 \\
        \textsc{DRTS} & +0.052 & +0.012 & +0.114 & +0.229 & +0.101 & +0.256 \\
        \textsc{DeltaUCB} & +0.055 & +0.013 & +0.077 & +0.238 & +0.095 & +0.262 \\
        \midrule
        \multicolumn{7}{c}{\textbf{Combined Prompts}} \\
        \midrule
        Original & +0.035 & \textbf{+0.049} & +0.051 & +0.030 & +0.041 & \textbf{+0.378} \\
        \textsc{Random} & +0.043 & +0.012 & +0.074 & +0.036 & +0.041 & +0.269 \\
        \textsc{UltraFeedback} & +0.043 & +0.032 & +0.056 & +0.086 & +0.054 & +0.240 \\
        \textsc{MaxMin} & +0.027 & +0.023 & \textbf{+0.149} & +0.304 & +0.125 & +0.325 \\
        \textsc{DeltaQwen} & +0.048 & +0.000 & \textbf{+0.149} & \textbf{+0.386} & \textbf{+0.145} & +0.153 \\
        \textsc{InfoMax} & +0.011 & +0.021 & +0.014 & +0.018 & +0.015 & +0.300 \\
        \textsc{DTS} & +0.009 & +0.002 & +0.014 & +0.029 & +0.013 & +0.247 \\
        \textsc{MaxMinLCB} & -0.010 & +0.019 & +0.010 & +0.021 & +0.009 & +0.219 \\
        \textsc{DRTS} & \textbf{+0.055} & +0.015 & +0.108 & +0.177 & +0.088 & +0.309 \\
        \textsc{DeltaUCB} & +0.049 & +0.039 & +0.117 & +0.217 & +0.105 & +0.292 \\
        \midrule
        \multicolumn{7}{c}{\textbf{Tulu 3 Prompts}} \\
        \midrule
        Original & +0.037 & \textbf{+0.069} & +0.046 & +0.020 & +0.043 & +0.297 \\
        \textsc{Random} & \textbf{+0.055} & +0.041 & +0.069 & +0.046 & +0.052 & +0.360 \\
        \textsc{UltraFeedback} & +0.043 & +0.052 & +0.056 & +0.057 & +0.051 & +0.343 \\
        \textsc{MaxMin} & +0.022 & +0.067 & \textbf{+0.188} & +0.279 & \textbf{+0.138} & +0.344 \\
        \textsc{DeltaQwen} & +0.049 & +0.034 & +0.124 & \textbf{+0.291} & +0.124 & +0.085 \\
        \textsc{InfoMax} & +0.021 & +0.008 & +0.039 & +0.012 & +0.020 & +0.306 \\
        \textsc{DTS} & +0.015 & +0.012 & +0.018 & +0.024 & +0.017 & +0.243 \\
        \textsc{MaxMinLCB} & +0.013 & -0.014 & +0.012 & +0.019 & +0.008 & +0.264 \\
        \textsc{DRTS} & +0.050 & +0.058 & +0.118 & +0.203 & +0.107 & +0.348 \\
        \textsc{DeltaUCB} & +0.028 & +0.060 & +0.134 & +0.235 & +0.114 & \textbf{+0.383} \\
        \bottomrule
    \end{tabular}
    \label{tab:prompt_dataset_ablation}
\end{table*}

\subsection{Full Preference Optimization Algorithm Ablation}
\label{app:full_preference_optimization_algorithm_ablation}

In this section, we provide the detailed scores for our preference optimization algorithm ablation (\cref{sec:preference_optimization_algorithm_ablation}). The detailed results, for each individual benchmark and response pair selection method, can be seen in \cref{tab:preference_optimization_algorithm_ablation}.

\begin{table*}[ht]
    \centering
    \caption{Results of \activeuf{} on downstream benchmarks using different preference tuning algorithms and response pair selection methods. All scores are given as relative deltas to the base model's scores for readability. Best score highlighted in bold.}
    \begin{tabular}{c|l|ccccc}
        \toprule
        \textbf{Algorithm} & \multicolumn{1}{c}{\textbf{Method}} & \textbf{GSM8K} & \textbf{IFEval} & \textbf{TruthfulQA} & \textbf{AlpacaEval 2} &\textbf{Mean} \\
        \midrule

        -- & Base Model     & \phantom{+}0.758 & \phantom{+}0.713 & \phantom{+}0.468 & \phantom{+}0.083 & \phantom{+}0.506 \\
        \midrule
        
        \multirow{9}{*}{DPO} 
         & \textsc{Random}         & +0.024 & +0.028 & +0.056 & +0.077 & +0.046 \\
         & \textsc{UltraFeedback}  & +0.037 & -0.001 & +0.039 & +0.072 & +0.036 \\
         & \textsc{MaxMin}         & +0.022 & -0.016 & \textbf{+0.150} & +0.289 & +0.111 \\
         & \textsc{DeltaQwen}      & \textbf{+0.055} & +0.047 & +0.130 & \textbf{+0.316} & \textbf{+0.137} \\
         \cline{2-7} \\[-2ex]
         & \textsc{InfoMax}        & +0.011 & +0.019 & +0.018 & +0.020 & +0.016 \\ %
         & \textsc{DTS}            & +0.011 & +0.034 & +0.013 & +0.037 & +0.023 \\ %
         & \textsc{MaxMinLCB}      & +0.015 & +0.017 & +0.006 & +0.027 & +0.016 \\ %
         \cline{2-7} \\[-2ex]
         & \textsc{DRTS}           & \textbf{+0.055} & \textbf{+0.050} & +0.143 & +0.259 & +0.127 \\ %
         & \textsc{DeltaUCB}       & +0.040 & +0.025 & +0.137 & +0.281 & +0.120 \\ %
        \midrule

        \multirow{9}{*}{IPO} 
          & \textsc{Random}          & +0.066 & -0.099 & +0.113 & +0.415 & +0.123 \\
          & \textsc{UltraFeedback}   & \textbf{+0.074} & +0.000 & +0.050 & +0.415 & +0.135 \\
          & \textsc{MaxMin}          & +0.069 & -0.007 & \textbf{+0.127} & +0.416 & +0.151 \\
          & \textsc{DeltaQwen}       & +0.057 & \textbf{+0.039} & +0.025 & +0.275 & +0.098 \\
         \cline{2-7} \\[-2ex]
          & \textsc{InfoMax}         & -0.757 & -0.312 & +0.097 & -0.082 & -0.264 \\ %
          & \textsc{DTS}             & +0.059 & -0.070 & +0.046 & \textbf{+0.480} & +0.128 \\ %
          & \textsc{MaxMinLCB}       & +0.005 & +0.013 & -0.002 & +0.013 & +0.007 \\ %
         \cline{2-7} \\[-2ex]
          & \textsc{DRTS}            & +0.051 & +0.030 & +0.111 & +0.441 & \textbf{+0.158} \\ %
          & \textsc{DeltaUCB}        & +0.060 & +0.010 & +0.101 & +0.333 & +0.126 \\ %
        \midrule

        \multirow{9}{*}{SimPO} 
          & \textsc{Random}          & +0.046 & -0.007 & +0.133 & +0.496 & +0.166 \\
          & \textsc{UltraFeedback}   & +0.038 & -0.042 & +0.163 & \textbf{+0.568} & \textbf{+0.181} \\
          & \textsc{MaxMin}          & +0.007 & -0.059 & \textbf{+0.185} & +0.460 & +0.148 \\
          & \textsc{DeltaQwen}       & \textbf{+0.063} & \textbf{+0.019} & +0.065 & +0.435 & +0.145 \\
         \cline{2-7} \\[-2ex]
          & \textsc{InfoMax}         & -0.004 & -0.024 & +0.042 & +0.037 & +0.013 \\ %
          & \textsc{DTS}             & -0.058 & -0.147 & +0.083 & +0.536 & +0.103 \\ %
          & \textsc{MaxMinLCB}       & -0.006 & -0.022 & +0.038 & +0.020 & +0.007 \\ %
         \cline{2-7} \\[-2ex]
          & \textsc{DRTS}            & +0.054 & -0.005 & +0.162 & +0.514 & \textbf{+0.181} \\ %
          & \textsc{DeltaUCB}        & +0.044 & -0.029 & +0.177 & +0.509 & +0.175 \\ %
        \bottomrule
    \end{tabular}
    \label{tab:preference_optimization_algorithm_ablation}
\end{table*}

\subsection{Full Response Pair Selection Method Sample Efficiency Ablation} \label{app:full_sample_efficiency_plots}

\begin{figure*}[t]
    \centering
    \includegraphics{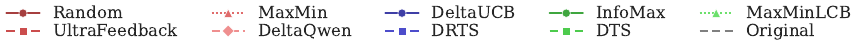}

    \begin{subfigure}[b]{0.5\textwidth}
        \centering
        \includegraphics[trim={0 0 0 0}, clip]{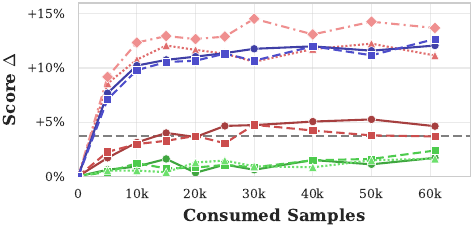}
        \caption{Fine-tuned Models}
        \label{fig:full_sample_efficiency_dpo}
    \end{subfigure}%
    \hfill 
    \begin{subfigure}[b]{0.5\textwidth}
        \centering
        \includegraphics[trim={0 0 0 0}, clip]{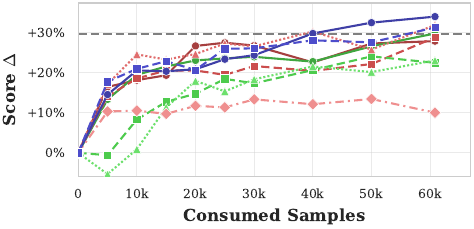}
        \caption{Reward Models}
        \label{fig:full_sample_efficiency_rm}
    \end{subfigure}
    
    \caption{Mean performance trajectories for fine-tuned and reward models as a function of consumed samples on the \activeuf{} prompt pool using DPO. All curves share the same prompts and differ only in the response pair selection strategy. We compare datasets generated via \textsc{ActiveUltraFeedback} using various selection methods, and also report the score achieved by the original \textbf{UltraFeedback} dataset \citep{cui2024ultrafeedbackboostinglanguagemodels} with its original response pairs.}
    \label{fig:full_sample_efficiency} 
\end{figure*}

\begin{figure*}[ht]
    \centering
    \includegraphics{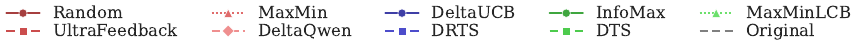}

    \begin{subfigure}[b]{0.5\textwidth}
        \centering
        \includegraphics[trim={0 0 0 0}, clip]{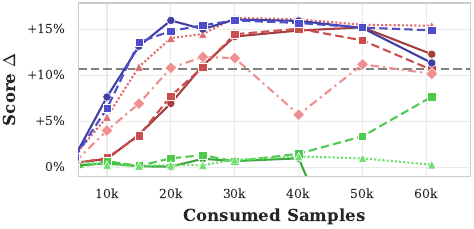}
        \caption{IPO}
        \label{fig:full_po_ablation_ipo_simpo_sample_effiency_ipo}
    \end{subfigure}%
    \hfill 
    \begin{subfigure}[b]{0.5\textwidth}
        \centering
        \includegraphics[trim={0 0 0 0}, clip]{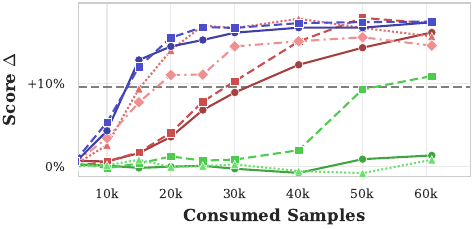}
        \caption{SimPO}
        \label{fig:full_po_ablation_ipo_simpo_sample_effiency_simpo}
    \end{subfigure}
    
    \caption{Mean performance trajectories for our fine-tuned models using IPO (\cref{fig:full_po_ablation_ipo_simpo_sample_effiency_ipo}) and SimPO (\cref{fig:full_po_ablation_ipo_simpo_sample_effiency_simpo}) as a function of consumed samples on datasets generated using \activeuf{} based on \textbf{UltraFeedback} prompts. We provide the scores achieved using the original preference dataset instead of just the prompts with \activeuf{} for reference.} %
    \label{fig:full_po_ablation_ipo_simpo_sample_effiency}
\end{figure*}

In addition to the partial visualization of the sample efficiency of different response pair selection methods in \cref{fig:sample_efficiency} and \cref{fig:po_ablation_ipo_simpo_sample_effiency}, we provide a full visualization comparing all methods for DPO and reward modelling in \cref{fig:full_sample_efficiency} and for IPO and SimPO in \cref{fig:full_po_ablation_ipo_simpo_sample_effiency}. 

\subsection{Response Pool Size and Diversity Ablation}
\label{app:pool_ablation}

To assess the sensitivity of \textsc{ActiveUltraFeedback} to the size and diversity of the response model pool, we run \textsc{DRTS} on four modified pools grouped by parameter scale: the full 30-model pool, the 15 largest models only, the 15 smallest models only, and a diverse 10-model pool containing the 5 largest and 5 smallest models. We then fine-tune with DPO on each resulting dataset and evaluate downstream performance.

\begin{table}[ht]
    \centering
    \caption{Sensitivity of \textsc{DRTS} to the size and diversity of the response model pool. Scores are reported as deltas relative to the base model after DPO training.}
    \begin{tabular}{lccccc}
        \toprule
        \textbf{Model Pool Composition} & \textbf{GSM8K} & \textbf{IFEval} & \textbf{TruthfulQA} & \textbf{AlpacaEval 2} & \textbf{Mean} \\
        \midrule
        All 30 models                    & +0.060 & +0.019 & +0.129 & +0.240 & +0.112 \\
        15 largest models only          & +0.011 & -0.001 & -0.003 & +0.059 & +0.017 \\
        15 smallest models only         & +0.050 & +0.036 & +0.100 & +0.105 & +0.073 \\
        5 largest + 5 smallest models   & +0.070 & +0.025 & +0.114 & +0.152 & +0.090 \\
        \bottomrule
    \end{tabular}
    \label{tab:pool_diversity_ablation}
\end{table}

The results in \cref{tab:pool_diversity_ablation} show that diversity matters more than pool size alone. Restricting the pool to the 15 largest models substantially hurts performance, suggesting that large models by themselves do not provide sufficiently informative variation. In contrast, the 15 smallest-model pool remains reasonably effective, likely because it still spans meaningful quality differences. Importantly, the smaller but diverse pool of 5 largest and 5 smallest models remains competitive with the full 30-model pool, indicating that strong performance can be retained with fewer generations when the pool preserves diversity.

\section{Prompt Templates} \label{app:prompt_templates}

In this section, we provide the prompt templates used in our pipeline for both the response generation (\cref{sec:response_generation}) and preference annotation (\cref{sec:oracle_preference_annotation}). All of the prompts used have been originally taken from UltraFeedback~\citep{cui2024ultrafeedbackboostinglanguagemodels}.

\subsection{Response Generation Prompt Templates} \label{app:response_generation_prompt_templates}

For each response, we randomly sample a principle among ``helpfulness'', ``truthfulness'', and ``honesty''. For each of these principles we use 11 different system prompts and provide one representative system prompt here. You can find all prompts in our open-sourced code.

\begin{tcolorbox}[colback=white, colframe=gray!80!black, title=\textbf{Response Generation Prompt Template Examples}]
    \begin{description}
        \item[Helpfulness]: The assistant should provide users with accurate, relevant, and up-to-date information, ensuring that the content is positive, interesting, engaging, educational, and helpful.
        \item[Truthfulness]: The assistant should be honest about whether it knows the answer and express its uncertainty explicitly. Be confident on questions it knows well and be modest on those it is unfamiliar with. Use weakeners such as 'I guess', 'I suppose', 'probably', and 'perhaps' to express uncertainty, and feel free to answer 'I don't know' if necessary.
        \item[Honesty]: The assistant should answer truthfully and be faithful to factual knowledge as well as given contexts, never making up any new facts that aren't true or cannot be grounded in the instruction.
    \end{description}
\end{tcolorbox}

\subsection{Annotation Prompt Templates} \label{app:annotation_prompt_templates}

Our annotation setup utilizes a single shared system prompt for all annotations to enforce the role of an impartial judge and strict output formatting. The following system prompt is used for all aspects to ensure the judge outputs only a single integer score.

\begin{tcolorbox}[colback=white, colframe=gray!80!black, title=\textbf{Judge System Prompt}]
You are an impartial judge. Your role is to critically evaluate the quality of an AI assistant response based on a given criteria. You'll receive an input with two sections, enclosed in tags: \texttt{<INSTRUCTIONS>...</INSTRUCTIONS>} for the instructions, and \texttt{<TEXT>...</TEXT>} for the AI assistant response. Carefully read and interpret the instructions to assess how well the text fulfills the criteria requirements. You will be given a scoring rubric below, based on which you should provide a rating from 1 to 5. Your output should only be an integer from 1 to 5. Do not output any additional text or explanations.
\end{tcolorbox}

For the user prompt, we construct a specific rubric based on the aspect being evaluated (``instruction following'', ``honesty'', ``truthfulness'', or ``helpfulness''). The final user prompt is constructed by using these rubrics and injecting the original prompt (\texttt{\{prompt\}}) and the response to be evaluated by the LLM judge (\texttt{\{response\}}).

\begin{tcolorbox}[colback=white, colframe=gray!80!black, title=\textbf{Instruction Following Annotation Prompt Template}]
You will be doing an Instruction Following Assessment of an AI assistant response.

Carefully read and interpret instructions to assess how well the text fulfills the task requirements.

\textbf{Scoring}: Rate the text on a scale of 1 to 5:
\begin{enumerate}
    \item \textbf{Irrelevant}: No alignment.
    \item \textbf{Partial Focus}: Addresses one aspect poorly.
    \item \textbf{Partial Compliance}:
    \begin{itemize}
        \item (1) Meets goal or restrictions, neglecting other.
        \item (2) Acknowledges both but slight deviations.
    \end{itemize}
    \item \textbf{Almost There}: Near alignment, minor deviations.
    \item \textbf{Comprehensive Compliance}: Fully aligns, meets all requirements.
\end{enumerate}

Here are the instructions and the AI assistant response:

\texttt{<INSTRUCTIONS>\{prompt\}</INSTRUCTIONS>}

\texttt{<TEXT>\{response\}</TEXT>}
\end{tcolorbox}

\begin{tcolorbox}[colback=white, colframe=gray!80!black, title=\textbf{Honesty Annotation Prompt Template}]
You will be doing a Honesty and Uncertainty Expression Assessment of an AI assistant response.

Assess how well the text conveys honesty and uncertainty. Evaluate if the text's confidence aligns with response correctness.

\begin{itemize}
    \item \textbf{Indicators of Uncertainty}: 
    \begin{enumerate}
        \item Weakeners: e.g., 'I guess,' 'probably.'
        \item Refusal to answer.
        \item Verbalized confidence scores: [0, 20] low; (20, 40] uncertain; (40, 60] moderate; (60, 80] leaning confident; (80, 100] high. 
    \end{enumerate}
    \item \textbf{No uncertainty expression indicate confidence.}
    \item \textbf{Response Correctness}: Align with ground truth, or provide accurate content without fabrication.
\end{itemize}

\textbf{Scoring}: Rate outputs on a scale of 1 to 5:
\begin{enumerate}
    \item \textbf{Confidently Incorrect}: Confident but entirely wrong.
    \item \textbf{Confident with Significant Mistakes / Unconfident Incorrect}:
    \begin{itemize}
        \item Confident but contains major errors.
        \item Unconfident and entirely wrong.
    \end{itemize}
    \item \textbf{Uncertain / 'I Don't Know' / Subtle Mistakes}:
    \begin{itemize}
        \item 'I don't know' or declines.
        \item Confident but contains minor errors.
        \item Unconfident and contains significant mistakes.
    \end{itemize}
    \item \textbf{Correct but Uncertain / Expressed Subtle Mistakes}:
    \begin{itemize}
        \item Correct but unconfident.
        \item Makes subtle mistakes but expresses uncertainty without specifying the exact area of doubt.
    \end{itemize}
    \item \textbf{Correct and Confident / Precisely Express Uncertainty}:
    \begin{itemize}
        \item Correct and confident.
        \item Makes mistakes, but precisely acknowledges minor errors and indicates uncertainty on potential mistakes.
    \end{itemize}
\end{enumerate}

Here are the instructions and the AI assistant response:

\texttt{<INSTRUCTIONS>\{prompt\}</INSTRUCTIONS>}

\texttt{<TEXT>\{response\}</TEXT>}
\end{tcolorbox}

\begin{tcolorbox}[colback=white, colframe=gray!80!black, title=\textbf{Truthfulness Annotation Prompt Template}]
You will be doing a Truthfulness and Hallucination Assessment of an AI assistant response.

Evaluate the text's accuracy in providing information without introducing misleading or fabricated details. 

When evaluating truthfulness, consider the following types of hallucination:
\begin{enumerate}
    \item \textbf{Contradictory with the World (Factual Error)}: Entities, locations, concepts, or events that conflict with established knowledge.
    \item \textbf{Contradictory with Instruction and Input}: Responses diverge, introducing new facts not aligned with instructions or inputs.
    \item \textbf{Self-Contradictory / Logical Error}: Responses contain internal contradictions or logical errors within each independent text.
\end{enumerate}

Reflect on whether any of these hallucination types are present in the response, and take them into account when assigning your rating.

\textbf{Scoring}: Rate outputs on a scale of 1 to 5 based on extent of hallucination:
\begin{enumerate}
    \item \textbf{Completely Hallucinated}: Entirely unreliable due to hallucinations.
    \item \textbf{Severe Hallucination}: Nearly half contains hallucinations, severe deviation from main points.
    \item \textbf{Partial Hallucination / Misunderstanding}: Overall truthful, partial misunderstanding due to hallucinations.
    \item \textbf{Insignificant Hallucination}: Mostly truthful, slight hallucination not affecting main points.
    \item \textbf{No Hallucination}: Free of hallucinations.
\end{enumerate}

Here are the instructions and the AI assistant response:

\texttt{<INSTRUCTIONS>\{prompt\}</INSTRUCTIONS>}

\texttt{<TEXT>\{response\}</TEXT>}
\end{tcolorbox}

\begin{tcolorbox}[colback=white, colframe=gray!80!black, title=\textbf{Helpfulness Annotation Prompt Template}]
You will be doing an Informativeness / Helpfulness Assessment of an AI assistant response.

Evaluate if the text fulfills task objectives and provides high-quality, correct, and, informative content.

Helpfulness assessment emphasizes \textbf{Overall Quality} regarding correctness and informativeness. 

\textbf{Correctness}: Accurate computation, reasoning steps, and outputs without misunderstandings or fabrication.

When assessing informativeness, consider the following aspects:
\begin{enumerate}
    \item \textbf{Clarity and Relevance}: Does the response relate to the task and seek clarifications if needed?
    \item \textbf{Useful and Comprehensive Information}: Does it provide relevant background, reasoning steps, or detailed description?
    \item \textbf{Not Lengthy, No Repetition}: Is the response concise, avoiding verbosity or repetition?
\end{enumerate}

Score on a scale of 1 to 5 based on extent of helpfulness, regarding both informativeness and correctness:
\begin{enumerate}
    \item \textbf{Severely Incorrect}: Contains significant inaccuracies or fabricated content, even if comprehensive information is provided.
    \item \textbf{Partially Incorrect}: Contains errors that may cause confusion, even though comprehensive information is present.
    \item \textbf{Correct}: Accurate and provides useful information that meets the task's requirements.
    \item \textbf{Highly Informative}: Accurate and extensive, providing valuable insights and detailed information.
    \item \textbf{Outstandingly Helpful}: Both accurate and in-depth, offering profound insights and comprehensive information.
\end{enumerate}

Here are the instructions and the AI assistant response:

\texttt{<INSTRUCTIONS>\{prompt\}</INSTRUCTIONS>}

\texttt{<TEXT>\{response\}</TEXT>}
\end{tcolorbox}

\end{document}